\documentclass[a4paper,fleqn]{cas-sc}

\usepackage{graphicx}
\usepackage{epstopdf}
\usepackage{caption}

\usepackage{amsmath}
\usepackage{algorithm}
\usepackage{algpseudocode}
\usepackage[numbers,sort&compress]{natbib}

\def\tsc#1{\csdef{#1}{\textsc{\lowercase{#1}}\xspace}}
\tsc{WGM}
\tsc{QE}

\begin{document}
\let\WriteBookmarks\relax
\def\floatpagepagefraction{1}
\def\textpagefraction{.001}
\let\printorcid\relax
\captionsetup[figure]{labelfont={bf},labelformat={default},labelsep=period,name={Fig.}}

\shorttitle{}    

\shortauthors{}  

\title [mode = title]{A novel Trunk Branch-net PINN for flow and heat transfer prediction in porous medium}  



%

\author[1,2]{Haoyun Xing}
\credit{Conceptualization, Formal analysis, Investigation, Methodology, Writing – original draft}
\author[3]{Kaiyan Jin}
\credit{Data curation, Methodology, Methodology}
\author[1,3]{Guice Yao}
\credit{Supervision, Project administration, Writing – review and editing}
\ead{yaoguice@buaa.edu.cn}
\cormark[1]
\author[1]{Jin Zhao}
\credit{Supervision, Writing – review and editing, Methodology, Funding acquisition}
\ead{jin.zhao@buaa.edu.cn}
\cormark[1]
\author[4]{Dichu Xu}
\credit{Project administration, Writing – review and editing}
\author[1,2,3,4]{Dongsheng Wen}
\credit{Funding acquisition, Resources, Writing – review and editing, Conceptualization}
\ead{d.wen@buaa.edu.cn}
\cormark[1]





\affiliation[1]{organization={ School of Aeronautic Science and Engineering, Beihang University, Beijing, 100191, China}
            }
\affiliation[2]{organization={ Institute of Thermodynamics, Technical University of Munich, Munich, 80333, Germany}
            }
\affiliation[3]{organization={Sino-French Engineer School /School of General Engineering, Beihang University, Beijing, 100191, China}
            }
\affiliation[4]{organization={Multiphase Flow and New Energy Technology Laboratory, Ningbo Institute of Technology, Beihang University, Ningbo, 315800, China}
            }

\cortext[1]{Corresponding authors}


\begin{abstract}
A novel Trunk-Branch (TB)-net physics-informed neural network (PINN) architecture is developed, which is a PINN-based method incorporating trunk and branch nets to capture both global and local features. The aim is to solve four main classes of problems: forward flow problem, forward heat transfer problem, inverse heat transfer problem, and transfer learning problem within the porous medium, which are notoriously complex that could not be handled by origin PINN. In the proposed TB-net PINN architecture, a Fully-connected Neural Network (FNN) is used as the trunk net, followed by separated FNNs as the branch nets with respect to outputs, and automatic differentiation is performed for partial derivatives of outputs with respect to inputs by considering various physical loss. The effectiveness and flexibility of the novel TB-net PINN architecture is demonstrated through a collection of forward problems, and transfer learning validates the feasibility of resource reuse. Combining with the superiority over traditional numerical methods in solving inverse problems, the proposed TB-net PINN shows its great potential for practical engineering applications.
\end{abstract}




\maketitle
\section{Introduction}\label{Section 1}
Flow and heat transfer in porous medium have attracted much attention recently, which have been widely explored in various fields, such as aerospace thermal control \cite{zhang2020research,uyanna2020thermal}, fuel cell \cite{wang2009experimental} and geothermal energy systems \cite{bringedal2016upscaling}. Corresponding researches have been conducted in terms of both experiments and numerical simulations. Taking aerospace thermal control as an example, most of the experimental studies can be broadly categorized into two classes. The first category is concentrated on revealing phenomenological characteristics of flow and heat transfer, including ice formation under high enthalpy environment \cite{shen2016experimental}, vapor blockage effect \cite{huang2017experimental}, and heat transfer or skin friction coefficients in different porous medium types \cite{jiang2004experimental,jiang2004experimental1}. The second category is focused on assessing cooling effectiveness, including the influences of the injection rates, thermal conductivities and particle diameters \cite{liu2010transpiration,liu2013experimental}. These experimental studies have promoted the optimization and improvement of porous medium structure parameters and heat transfer efficiency. The non-transparent basis material and the complex micro-structure of the porous media, however, have prevented the detailed measurements of velocity, pressure and temperature distributions inside the structure. Mostly only surface quantities such as temperature could be captured by infrared thermal image systems (ITIS) readily.

As a complementary tool to the experimental measurements, numerical simulation reveals its importance of forecasting the whole spatio-temporal distributions of interested variables. According to the simulation scale, numerical methods can be classified into two forms, i.e., pore-scale and Darcy-scale. Pore-scale methods include the lattice Boltzmann method \cite{fei2023lattice,fu2024pressure}, the pore network model \cite{wang2023numerical,le2016pore} and the volume of fluid method \cite{salimi2024volume, patel2019effect}. For pore-scale methods, geometry and topology of the medium should be represented and sophisticated grid processing is necessary. It turns out that such methods face intractable difficulties in large-scale flow and heat transfer problems due to the computational cost. On the contrary, Darcy-scale methods, consider porous media as continuous cavity structure with extra permeability by using the concept of representative element volume (REV). Traditionally, the so-called separated flow model (SFM) \cite{scheidegger1957physics, bear2013dynamics} has been employed in Darcy-scale. In this model, different phases are described by their respective conservation laws, leading to large number of conservative equations to be solved. As such, the SFM is generally inconvenient for practical problems, especially in multi-dimensional situations. Alternatively, the two-phase mixture model (TPMM) \cite{chao1993two} has been developed by considering different phases as the constituents of a multicomponent mixture, which largely reduces the number of differential equations. By introducing the mixture enthalpy concept, the numerical convergence of TPMM is guaranteed. Notwithstanding the widely adoption of TPMM in engineering applications \cite{he2013modeling, su2019numerical}, it suffers the problems of mesh generation for complicated geometries and the difficulties of incorporating available data into the numerical methods seamlessly.

Recent advancements in machine learning \cite{jordan2015machine, lecun2015deep} and data science \cite{marx2013big} have promoted significant progress in computer vision (CV) \cite{lecun1989backpropagation}, natural language processing (NLP) \cite{arulkumaran2017deep}, active control \cite{vaswani2017attention} and feature distilling \cite{xing2022using}. Despite the remarkable success, machine learning has encountered many obstacles in scientific computing due to the intricate physics. The first glimpses of promise traced back to 1990’s \cite{psichogios1992hybrid,lagaris1998artificial}, the physical laws were encoded into the loss function of neural networks (NNs), and corresponding partial differential equations (PDE) could be solved by this way. However, owing to the limitation of computational resources, further exploration has been stagnated for decades. With recent improvement in hardware and software, including graphic processing units (GPUs), tensor processing units (TPUs) and open-source libraries, the concept has been revisited for more challenging problems, and re-named as physics-informed neural network (PINN) \cite{raissi2019physics}. PINN harnesses Automatic Differentiation (AD) \cite{baydin2018automatic}, which differentiates NNs with respect to their inputs and parameters, and this make it a mesh-free approach compared with conventional numerical methods. By leveraging the capability as universal function approximators of NNs, there is empirical evidence articulated that, PINN is capable of achieving good prediction accuracy when relative PDE and solution are well-posed and unique, respectively \cite{lu2021deepxde}. In addition to solving forward problems with initial and boundary conditions \cite{jin2021nsfnets,gao2021phygeonet,yang2021b,biswas2023three}, PINN could tackle inverse problems with extra given data smoothly \cite{raissi2020hidden, jagtap2022physics} benefitted from the NNs structure.

In the fluid dynamics community, there has been a few explorations about flow and heat transfer problems in porous media using PINNs-like methods. For instance, Hanna et al. \cite{hanna2022residual} employed a separated NN with a residual-based point collocation strategy to tackle the two-phase flow forward problems with labeled data, however, the predictive outcome exhibited marked deviation from the exact solution. Wang et al. \cite{wang2020deep} proposed a theory-guided neural network (TgNN) to solve subsurface flows, it looks, however, more like a data-driven approach due to the abundance of labeled data when compared to other PDE solvers. In a later work, Wang et al. \cite{wang2021efficient} transitioned TgNN from fully-connected neural network (FNN) to a convolutional neural network (CNN) to investigate single-phase flow problems in a subsurface reservoir, incorporating discretized governing equations with finite difference (FD) into the loss function. Subsequently, Zhang et al. leveraged the physics-informed convolutional neural network (PICNN) to simulate both single-phase \cite{zhang2022physics} and two-phase \cite{zhang2023physics} Darcy flows in the absence of labeled data, and the finite volume discretization scheme is adopted to build the loss function. The adoption of CNN can expedite the inference speed, it lacks however the flexibility of FNN when confronted with irregular computational domains, reducing the reliability of prediction accuracy. Consequently, it is imperative to develop novel NN architectures and effective loss function configurations to handle such elaborate issues. It shall be noted that these prior studies have been focused on flow problems only, and the research on convective heat transfer in porous media is rarely reported. Recently, Amini et al. \cite{amini2023inverse} conducted reverse modeling of non-isothermal poromechanics via PINN, and the model parameters could be recognized under a similar framework as the forward problem, similar to their forward problems for thermal-hydro-mechanical processes with labeled data \cite{amini2022physics} that achieved limited accuracy. It is important to note that there is still no machine learning work in solving the forward problems of heat transfer in porous media without labeled data, especially considering local thermal non-equilibrium (LTNE) effect \cite{amiri1994analysis}, which employs a more realistic treatment of heat transfer in porous media compared with the local thermal equilibrium (LTE). In addition, the incorporation of transfer learning \cite{pan2009survey}, which could save computational resources and improve prediction precision \cite{radford2018improving, jumper2021highly} greatly due to the strong representation ability of NNs, is yet to be introduced.

Aiming to address above limitations, this work develops a pioneering trunk-branch (TB)-net architecture PINN for flow and heat transfer in porous media that is capable for solving four main problems: forward flow, forward heat transfer, inverse heat transfer and transfer learning. As a benchmarking study, the flow and heat transfer model in the porous media considering LTNE effect is employed. In this novel architecture, the solving process of flow and heat transfer are separated because of the numerous and complex loss items. Flow problem is firstly settled using corresponding branch nets, and with the obtained model as the default, heat transfer problem is solved through the counterparts via a step-wise strategy. Subsequently, inverse heat transfer problems with scattered temperature data distributed on the outlet boundary are investigated by means of adding data loss item and removing heat flux conditions. Finally, transfer learning technique is leveraged to work out flow and heat transfer problems under different porosities, external pressure and heat flux boundary conditions, which taking advantage of existing models. The rest of the paper is organized as follows. In Section \ref{Section 2}, the model formulation of the flow and heat transfer problems in porous media is introduced, and the TB-net PINN architecture is described afterwards. In Section \ref{Section 3}, computational cases including forward problems, inverse problems, and transfer learning attempts are discussed, along with the introduction of order of magnitude (OOM) and importance analysis for weight design. Finally, the main results are summarized in Section \ref{Section 4} with an outlook for the future work.

\section{Model formulation and TB-net PINN architecture}\label{Section 2}
\subsection{Model formulation}\label{Section 2.1}

\begin{figure}[htbp]
  \centering
    \includegraphics[scale=0.7]{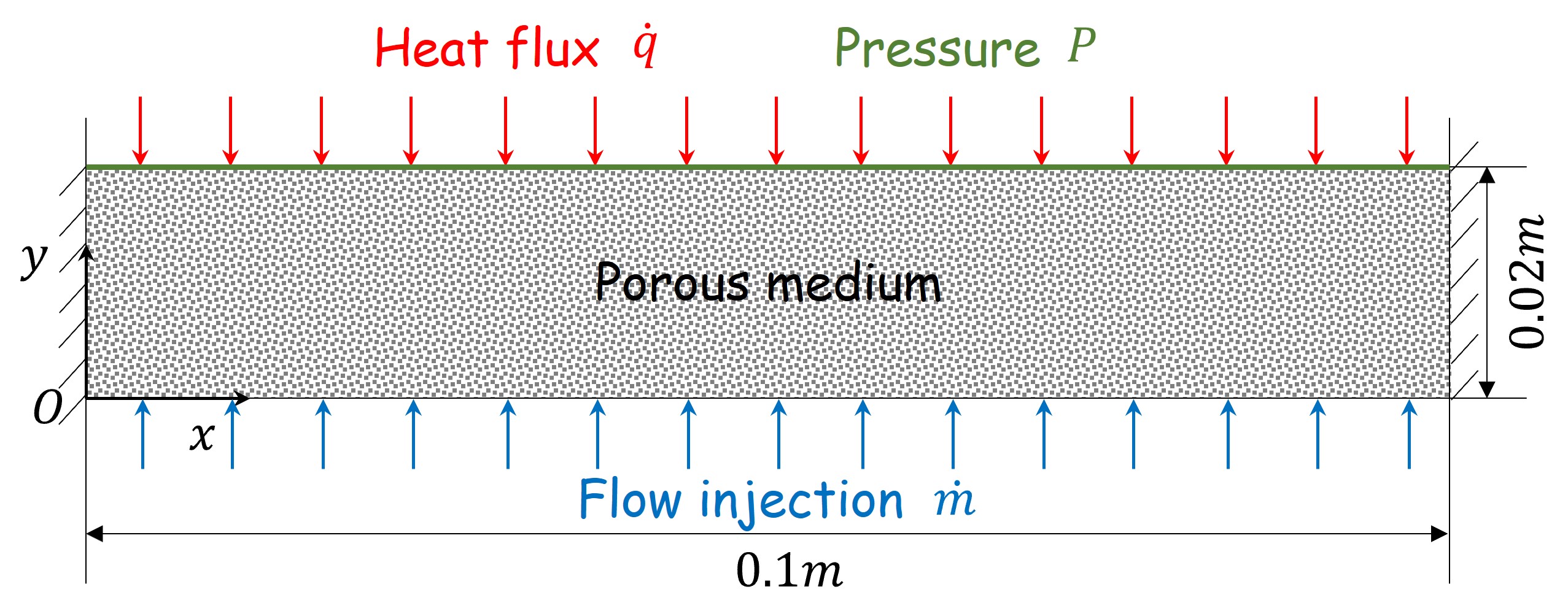}
    \caption{Physical model of the flow and heat transfer in the porous medium employed in this work.}\label{fig1}
\end{figure}

The steady-state flow and heat transfer model in the porous medium considering LTNE effect is employed, which involving the realistic non-equilibrium heat transfer phenomenon compared with the traditional local thermal equilibrium (LTE) \cite{hunt1988non}, i.e., by considering heat convection between solid skeleton and fluid. Considering a two-dimensional porous physical model with size 0.1 m $\times$ 0.02 m described in Fig. \ref{fig1}, the general form of the mathematical model for incompressible, viscid flow and heat transfer in a porous medium can be showcased by \cite{jin2024modified}:

\begin{itemize}
\item Continuum equation:
\begin{subequations}
\begin{align}
        u_x + v_y = 0, \label{Eq.1a} 
    \end{align}
\item Momentum equations:
\begin{align}
        \frac{\rho}{\varepsilon}((u^2)_x+(uv)_y)=-p_x-\frac{\mu}{K}u, \label{Eq.1b} 
        \\
        \frac{\rho}{\varepsilon}((vu)_x+(v^2)_y)=-p_y-\frac{\mu}{K}v, \label{Eq.1c}
    \end{align}
\item Enthalpy equations:
\begin{align}
        & Fluid:\rho((uh_k)_x+(vh_k)y)=\frac{\varepsilon k_f}{c_{pf}}((h_k)_{xx}+(h_k)_{yy})+h_{sf}\alpha_{sf}(T_s-T_f), \label{Eq.1d}& \\
        &Solid\ skeleton: (1-\varepsilon)k_s((T_s)_{xx}+(T_s)_{yy})=h_{sf}\alpha_{sf}(T_s-T_f), \label{Eq.1e}&
    \end{align}
\end{subequations}

\end{itemize}
where, $\textit{u}$ and $\textit{v}$ represent horizontal and vertical velocity of the fluid respectively; $\rho$ is the density of fluid; $\textit{p}$ denotes the pressure; the subscripts $\textit{x}$ and $\textit{y}$ denote partial differentiation in space; $\varepsilon$, $\textit{K}$, and $k_s$ specify the porosity, absolute permeability, and thermal conductivity of solid skeleton, respectively; $T_s$ corresponds to the temperature of solid. Besides, $\alpha_{sf}$ refers to the specific surface area of the skeleton and $h_{sf}$ is the convective heat transfer coefficient suggested as \cite{wakao1979effect}
\begin{equation}
  h_{sf}=k_f(2.0+1.1Pr_f^{0.33}Re_f^{0.6})/d_p, \label{Eq.2}
  \end{equation}
where $Pr_f$ and $Re_f$ represent Prandtl and Reynolds number of fluid, respectively, and $d_p$ is the particle diameter. For the physical properties of fluid, $k_f$ represents thermal conductivity, $c_{pf}$ is specific heat capacity, and $T_f$ denotes temperature. It is worth emphasizing that, by invoking $h_k$, i.e., the kinetic enthalpy of the liquid, the representation of the formulation above is simplified substantially.

Before embedding Eqs.(\ref{Eq.1a}-\ref{Eq.1e}) into TB-net PINN architecture, the order of magnitude (OOM) analysis is carried out: $u$, $v$, $p$, $h_k$ and $T_s$ are variables concerned and would be placed on the output layer of NNs; in the meantime, spatial coordinates $x$ and $y$ would be placed on the input layer. 
It is notable that the OOM of primary variables vary dramatically, and training such a model with striking difference in OOM of outputs and inputs shows impossible. To resolve this issue, the governing equations are further non-dimensionalized by using characteristic parameters as follows:
\begin{subequations}
\begin{align}
        &\widetilde{v}=\frac{v}{V}, \label{Eq.3a}& \\
        &\widetilde{u}=\frac{u}{V},& \label{Eq.3b} \\
        &\widetilde{p}=\frac{p}{P},& \label{Eq.3c}
    \\
        & \widetilde{h_k}=\frac{T_fc_{pf}}{TC_P}=\frac{h_k}{TC_P},& \label{Eq.3d}\\
        & \widetilde{T_s}=\frac{T_s}{T},& \label{Eq.3e} \\
        & \widetilde{x}=\frac{x}{L},& \label{Eq.3f} \\
\intertext{and}
        & \widetilde{y}=\frac{y}{L}.& \label{Eq.3g}
    \end{align}
\end{subequations}
where,  $\widetilde{u}$, $\widetilde{v}$, $\widetilde{p}$, $\widetilde{h_k}$, $\widetilde{T_s}$, $\widetilde{x}$ and $\widetilde{y}$ denote the non-dimensional variables transformed from $u$, $v$, $p$, $h_k$, $T_s$, $x$ and $y$; while, $V$, $L$, $P$, $T$ and $C_P$ represent the characteristic velocity, length, pressure, temperature and specific heat capacity, respectively. It is worth noting that the OOM of non-dimensional variables would be $O(10^0)$ when selecting characteristic parameters reasonably. In the cases discussed in this work, the liquid is injected along the vertical direction, hence the absolute value of $u$ could be neglected. The non-dimensional governing equations for flow and heat transfer in the porous medium considering LTNE are employed for NNs training:
\begin{itemize}
\item Continuum equation:
\begin{subequations}
\begin{align}
        \widetilde{u}_{\widetilde{x}} + \widetilde{v}_{\widetilde{y}} = 0, \label{Eq.4a} 
    \end{align}
\item Momentum equations:
\begin{align}
        \frac{\rho}{\varepsilon}\frac{V^2}{L}((\widetilde{u}^2)_{\widetilde{x}} + (\widetilde{u}\widetilde{v})_{\widetilde{y}})=-\frac{P}{L}\widetilde{p}_{\widetilde{x}}-\frac{\mu}{K}V\widetilde{u}, \label{Eq.4b} 
        \\
        \frac{\rho}{\varepsilon}\frac{V^2}{L}((\widetilde{v}\widetilde{u})_{\widetilde{x}} + (\widetilde{v}^2)_{\widetilde{y}})=-\frac{P}{L}\widetilde{p}_{\widetilde{y}}-\frac{\mu}{K}V\widetilde{v}, \label{Eq.4c}
    \end{align}
\item Enthalpy equations:
\begin{align}
        & Fluid:\rho\frac{VTC_P}{L}((\widetilde{u}\widetilde{h_k})_{\widetilde{x}}+(\widetilde{v}\widetilde{h_k})_{\widetilde{y}})=\frac{\varepsilon k_f}{c_{pf}}\frac{TC_P}{L^2}((\widetilde{h_k})_{\widetilde{x}\widetilde{x}}+(\widetilde{h_k})_{\widetilde{y}\widetilde{y}})+h_{sf}\alpha_{sf}T(\widetilde{T_s}-\widetilde{h_k}), \label{Eq.4d}& \\
\intertext{and}
        &Solid\ skeleton: (1-\varepsilon)k_s\frac{T}{L^2}((\widetilde{T_s})_{\widetilde{x}\widetilde{x}}+(\widetilde{T_s})_{\widetilde{y}\widetilde{y}})=h_{sf}\alpha_{sf}T(\widetilde{T_s}-\widetilde{h_k}). \label{Eq.4e}&
    \end{align}
\end{subequations}

\end{itemize}
Besides the governing equations mentioned above, boundary conditions in the non-dimensional form are as follows:
\begin{itemize}
\item Inlet boundary:
\begin{subequations}
\begin{align}
        &\widetilde{v}= 1,& \label{Eq.5a} \\
        &\widetilde{u}=0,& \label{Eq.5b} \\
        &(1-\varepsilon)k_s\frac{T}{L}\widetilde{T_s}_{\widetilde{y}}=h_iT(\widetilde{T_s}-1),& \label{Eq.5c} \\
        &h_iT(\widetilde{T_s}-1)=\rho VTC_P(\widetilde{h_k}-1), & \label{Eq.5d}
    \end{align}
\item Outlet boundary:
\begin{align}
        &\widetilde{p}= 1,& \label{Eq.5e}
        \\
        &(1-\varepsilon)k_s\frac{T}{L}\widetilde{T_s}_{\widetilde{y}}=\dot q,& \label{Eq.5f}\\
        &\widetilde{h_k}_{\widetilde{y}}= 0,& \label{Eq.5g}
    \end{align}
\item Wall boundary:
\begin{align}
        &\widetilde{u}= 0,& \label{Eq.5h} \\
        &\widetilde{v}= 0,& \label{Eq.5i} \\
        &\widetilde{T_s}_{\widetilde{x}}=0,& \label{Eq.5j} \\
        &\widetilde{h_k}_{\widetilde{y}}=0.& \label{Eq.5k}
    \end{align}
\end{subequations}

\end{itemize}
Note that $h_i$ denotes convection heat transfer coefficient of the inlet fluid, and is defined as $h_i=0.664{Pr_i}^{\frac{1}{3}}{Re_i}^{\frac{1}{2}}$, where, $Pr_i$ and $Re_i$ represent Prandtl and Reynolds number of the inlet fluid, respectively, and $\dot{q}$ corresponds to the heat flux applied to the outlet boundary.

\subsection{TB-net PINN architecture}\label{Section 2.2}
\begin{figure}[htbp]
  \centering
    \includegraphics[scale=0.39]{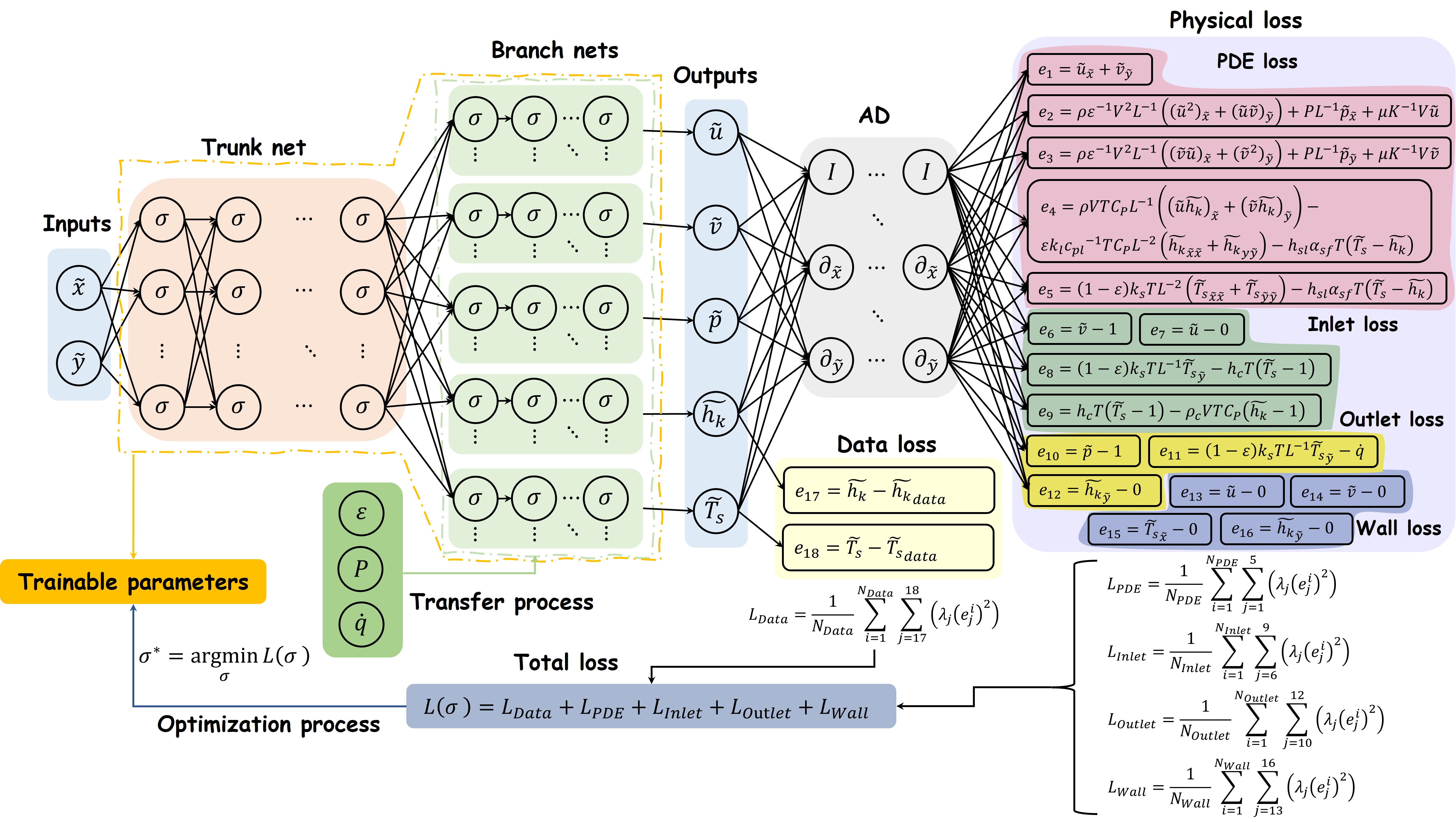}
    \caption{Schematic of the TB-net PINN architecture: Inputs: The non-dimensional spatial coordinates. Trunk net: A fully-connected neural network (FNN). Branch nets: Separated FNNs with respect to outputs. The trunk net and branch nets compose the TB-net. Outputs: The non-dimensional primary variables. AD: Automatic differentiation for partial derivatives of outputs with respect to inputs. Physical loss: The loss function includes five parts, i.e., PDE loss, inlet loss, outlet loss, wall loss, and data loss. The training parameters consist of weights and biases of trunk and branch nets. $\varepsilon$, $P$, and $\dot q$ are chosen factors for transfer learning process.}\label{fig2}
\end{figure}

A novel Trunk-Branch (TB)-net PINN is developed to tackle the flow and heat transfer problem in the porous media with LTNE effect described in Section \ref{Section 2.1}, as shown in Fig. \ref{fig2}. The proposed TB-net PINN cannot only recognize global and local features but also circumvent the challenge faced by vanilla PINN in distinguishing multi-physical fields solely through the last hidden layer during the solution process. Furthermore, flexibility and efficiency are also hallmarks of TB-net PINN. 

The main component of the architecture could be expressed as:
\begin{equation}(\widetilde{u},\widetilde{v},\widetilde{p},\widetilde{h_k},\widetilde{T_s})=\mathcal{T}\mathcal{B}_{NN}(\widetilde{x},\widetilde{y},\sigma_\mathcal{T},\sigma_\mathcal{B}), \label{Eq.6}
\end{equation}

where, $\sigma_\mathcal{T}$ and $\sigma_\mathcal{B}$ denote trainable parameters corresponding to trunk and branch nets, respectively. The trainable parameters can be learned by minimizing the loss function
\begin{equation}
\mathcal{L}(\sigma)=\mathcal{L}_{Data}+\mathcal{L}_{PDE}+\mathcal{L}_{Inlet}+\mathcal{L}_{Outlet}+\mathcal{L}_{Wall}, \label{Eq.7}
\end{equation}

where, $\mathcal{L}_{Data}$ is the data loss, $\mathcal{L}_{PDE}$ denotes the PDE loss, $\mathcal{L}_{Inlet}$ represents the loss of inlet condition, $\mathcal{L}_{Outlet}$ specifies the loss of outlet condition, and $\mathcal{L}_{Wall}$ corresponds to the loss of wall condition. Specifically,
\begin{subequations}
\begin{align}
    &\mathcal{L}_{PDE}=\frac{1}{N_{PDE}}\sum_{i=1}^{N_{PDE}}\sum_{j=1}^5(\lambda_j(e_j^i)^2),& \label{Eq.8a}\\
    &\mathcal{L}_{Inlet}=\frac{1}{N_{Inlet}}\sum_{i=1}^{N_{Inlet}}\sum_{j=6}^9(\lambda_j(e_j^i)^2),& \label{Eq.8b}\\
    &\mathcal{L}_{Outlet}=\frac{1}{N_{Outlet}}\sum_{i=1}^{N_{Outlet}}\sum_{j=10}^{12}(\lambda_j(e_j^i)^2),& \label{Eq.8c}\\
    &\mathcal{L}_{Outlet}=\frac{1}{N_{Wall}}\sum_{i=1}^{N_{Wall}}\sum_{j=13}^{16}(\lambda_j(e_j^i)^2),& \label{Eq.8d}\\
\intertext{and}
&\mathcal{L}_{Data}=\frac{1}{N_{Data}}\sum_{i=1}^{N_{Data}}\sum_{j=17}^{18}(\lambda_j(e_j^i)^2),& \label{Eq.8e}
    \end{align}
\end{subequations}
where, $N_{PDE}$, $N_{Inlet}$, $N_{Outlet}$, and $N_{Wall}$ specify the number of collocation points in the computational domain, on the inlet boundary, on the outlet boundary, and on the wall boundary, respectively, while $N_{Data}$ denote the number of extra training data. $\{e_j\}_{j=1}^{18}$ are loss terms transformed from Eqs.(\ref{Eq.4a}-\ref{Eq.4e}) and (\ref{Eq.5a}-\ref{Eq.5k}), which represent the constraint of physical laws, and could be written as:
\begin{subequations}
\begin{align}
    &e_1=\widetilde{u}_{\widetilde{x}} + \widetilde{v}_{\widetilde{y}},& \label{Eq.9a}\\
    &e_2=\frac{\rho}{\varepsilon}\frac{V^2}{L}((\widetilde{u}^2)_{\widetilde{x}} + (\widetilde{u}\widetilde{v})_{\widetilde{y}})+\frac{P}{L}\widetilde{p}_{\widetilde{x}}+\frac{\mu}{K}V\widetilde{u},& \label{Eq.9b}\\
    &e_3=\frac{\rho}{\varepsilon}\frac{V^2}{L}((\widetilde{v}\widetilde{u})_{\widetilde{x}} + (\widetilde{v}^2)_{\widetilde{y}})+\frac{P}{L}\widetilde{p}_{\widetilde{y}}+\frac{\mu}{K}V\widetilde{v},& \label{Eq.9c}\\
    &e_4=\rho\frac{VTC_P}{L}((\widetilde{u}\widetilde{h_k})_{\widetilde{x}}+(\widetilde{v}\widetilde{h_k})_{\widetilde{y}})-\frac{\varepsilon k_f}{c_{pf}}\frac{TC_P}{L^2}((\widetilde{h_k})_{\widetilde{x}\widetilde{x}}-(\widetilde{h_k})_{\widetilde{y}\widetilde{y}})+h_{sf}\alpha_{sf}T(\widetilde{T_s}-\widetilde{h_k}),& \label{Eq.9d}\\
    &e_5=(1-\varepsilon)k_s\frac{T}{L^2}((\widetilde{T_s})_{\widetilde{x}\widetilde{x}}+(\widetilde{T_s})_{\widetilde{y}\widetilde{y}})-h_{sf}\alpha_{sf}T(\widetilde{T_s}-\widetilde{h_k}),& \label{Eq.9e}\\
    &e_6=\widetilde{v}-1,& \label{Eq.9f}\\
    &e_7=\widetilde{u}-0,& \label{Eq.9g}\\
    &e_8=(1-\varepsilon)k_s\frac{T}{L}\widetilde{T_s}_{\widetilde{y}}-h_iT(\widetilde{T_s}-1),& \label{Eq.9h}\\
    &e_9=h_iT(\widetilde{T_s}-1)-\rho VTC_P(\widetilde{h_k}-1),& \label{Eq.9i}\\
    &e_{10}=\widetilde{p}-1,& \label{Eq.9j}\\
    &e_{11}=(1-\varepsilon)k_s\frac{T}{L}\widetilde{T_s}_{\widetilde{y}}-\dot q,& \label{Eq.9k}\\
    &e_{12}=\widetilde{h_k}_{\widetilde{y}}-0,& \label{Eq.9l}\\
    &e_{13}=\widetilde{u}-0,& \label{Eq.9m}\\
    &e_{14}=\widetilde{v}-0,& \label{Eq.9n}\\
    &e_{15}=\widetilde{T_s}_{\widetilde{x}}-0,& \label{Eq.9o}\\
    &e_{16}=\widetilde{h_k}_{\widetilde{y}}-0,& \label{Eq.9p}\\
    &e_{17}=\widetilde{h_k}-\widetilde{h_k}_{data},& \label{Eq.9q}\\
\intertext{and}
    &e_{18}=\widetilde{T_s}-\widetilde{T_s}_{data},& \label{Eq.9r}\
    \end{align}
\end{subequations}

Note that the $\widetilde{h_k}_{data}$ and $\widetilde{T_s}_{data}$ are extra training data related to non-dimensional fluid and solid temperature, respectively. Meanwhile, $\{\lambda_j\}_{i=1}^{18}$ specify the weight parameters used to balance each loss term, which could be tuned artificially. In a nutshell, $\mathcal{L}_{PDE}$ enforces the governing equations being satisfied on the collocation points in the computational domain, $\mathcal{L}_{Inlet}$, $\mathcal{L}_{Outlet}$, and $\mathcal{L}_{Wall}$ ensure a series of boundary conditions, and $\mathcal{L}_{Data}$ penalizes the difference between the outputs of $\mathcal{T}\mathcal{B}_{NN}$ and the given additional training data.

The concrete instantiation of Eq.(\ref{Eq.7}) consists of numerous partial derivative calculations, which would be addressed by using automatic differentiation (AD). In general, AD is implemented with the aid of chain rule and back propagation, and compared with finite difference or other numerical approximations, it avoids truncation error and the results could be more precise. As shown in Fig. \ref{fig2}, the AD part manages the partial derivatives of outputs with respect to inputs. Finally, the training parameters are optimized to achieve the state described by
\begin{equation}
\sigma^*=\mathop{argmin}_\sigma\mathcal{L}(\sigma), \label{Eq.10}
\end{equation}

In recognition of the fact that the $\mathcal{L}(\sigma)$ is highly nonlinear and nonconvex pertaining to $\sigma$, gradient-based methods are considered to search the optimal solutions. Since that stochastic gradient descent-based Adam method \cite{kingma2014adam} and quasi-Newton-based L-BFGS method \cite{liu1989limited} represent the cutting-edge technologies among gradient-based methods, these two optimizers are further employed to train the TB-net PINN architecture.

\section{Predictions of flow and heat transfer in porous medium}\label{Section 3}
In this section, the forward problems are firstly discussed to demonstrate the efficiency and effectiveness of the architecture compared with vanilla PINN in Section \ref{Section 3.1}. To illustrate the flexibility of the architecture is superior to the traditional numerical methods, several inverse and transfer learning problems are conducted in Section \ref{Section 3.2} and \ref{Section 3.3}, respectively.
In order to evaluate the performance of the proposed methodology, high-resolution test datasets are indispensable. In this context, finite volume method (FVM) is utilized for the flow simulation, starting with the generation of a structured grid with node number 400 $\times$ 500 in spatial coordinates. The second-order upwind scheme is adopted for the spatial discretization, and the Pressure-Implicit with Splitting of Operators (PISO) method is applied to the pressure-velocity coupling iteration. For the heat transfer simulation, one only needs to discretize and solve the energy equations based on the flow results. The reliability of the data generation method has been verified \cite{jin2024modified}, and the generated dataset would serve as exact values for validating prediction performance or extra labeled data.
\subsection{Forward problems}\label{Section 3.1}
In all cases pertaining to forward problems, the data loss is removed, i.e., without labeled data, and only physical loss remains. In other words, $\mathcal{L}_{Data}$ wouldn't be involved in the loss function. In this sense, this subsection straightly assesses the capability of the NN architecture and the rationality of the design of the inputs, outputs and loss function.
\subsubsection{Flow problem}\label{Section 3.1.1}
The investigations shown here aim to emphasize the ability of the proposed method to handle multi-physics solutions with different OOM and maintain the stability of the training process.

The material employed is the sintered particle porous media with properties shown in Table \ref{tbl1}, and the geometry size is 0.1 m $\times$ 0.02 m as described in Fig. \ref{fig1}. The liquid water is adopted as the fluid, and the relevant properties are given in Table \ref{tbl2}. The wall boundary is considered as no slip condition, and the outlet boundary is pressure condition with approximate standard atmosphere. The liquid is injected along the vertical direction in this work with various inlet mass flux ($\dot m=\rho v$), and three cases are carried out, i.e., $case$ A, B and C with $\dot m$ = 0.1, 0.5 and 1.0 kg/($\mathrm{m^2s}$), respectively.
\begin{table*}[pos=!htbp,width=0.7\textwidth]
\caption{Physical properties of the porous media employed in this work.}\label{tbl1}
\begin{tabular*}{\tblwidth}{@{}LLL@{}}
\toprule
 \textbf{Property} & \textbf{Value} & \textbf{Unit} \\ 
\midrule
 specific heat capacity & 4000 & J/(kg$\cdot$K) \\
 thermal conductivity & 30 & W/(m$\cdot$K) \\
 porosity & 0.3 & /\\
 particle diameter & 1 $\times$ $10^{-4}$ & m \\
 permeability & 3.67 $\times$ $10^{-12}$ & $\mathrm{m^2}$ \\
\bottomrule
\end{tabular*}
\end{table*}

\begin{table*}[pos=!htbp,width=0.7\textwidth]
\caption{Physical properties of the liquid water adopted in this work.}\label{tbl2}
\begin{tabular*}{\tblwidth}{@{}LLL@{}}
\toprule
 \textbf{Property} & \textbf{Value} & \textbf{Unit} \\ 
\midrule
 density & 960 & kg/($\mathrm{m^3}$) \\
 specific heat capacity & 4217 & J/(kg$\cdot$K) \\
 dynamic viscosity & 2.41 $\times$ $10^{\frac{247.8}{T_f-140}}$ $\times$ $10^{-5}$ & kg/(m$\cdot$s)\\
 thermal conductivity & 0.68 & W/(m$\cdot$K) \\
 Prandtl number & 0.149 $\times$ $10^{\frac{247.8}{T_f-140}}$ & / \\
\bottomrule
\end{tabular*}
\end{table*}

\textbf{Case A}. In this case, mass flux is set as $\dot m$ = 0.1 kg/($\mathrm{m^2}$s), and with the aid of characteristic parameters \textit{V} = 0.1/$\rho$, P = 1.0 $\times$ $10^5$ Pa and \textit{L} = 0.1 m, the dimensional boundary conditions would be converted into Eqs.(\ref{Eq.5a}-\ref{Eq.5b}), (\ref{Eq.5e}), and (\ref{Eq.5h}-\ref{Eq.5i}). In a nutshell, the loss function for flow problems includes loss terms $e_1$$\sim$$e_3$, $e_6$$\sim$$e_7$, $e_{10}$ and $e_{13}$$\sim$$e_{14}$. The corresponding computational domain is transformed into:
\begin{equation}
\begin{aligned}
&x/L=\widetilde{x}\in[0,1],\\
&y/L=\widetilde{y}\in[0,0.2]\label{Eq.11}
\end{aligned}
\end{equation}

It is worth emphasizing that the performance of PINNs-like framework would be susceptible to the weight parameters corresponding to the loss terms, hence the weights are assigned by OOM and importance analysis to guarantee the capability of TB-net PINN. Specifically, the OOM of the loss terms is $e_1$ $\sim$ \textit{O}($10^0$), $e_2$ $\sim$ \textit{O}($10^4$), $e_3$ $\sim$ \textit{O}($10^4$), $e_6$ $\sim$ \textit{O}($10^0$), $e_7$ $\sim$ \textit{O}($10^0$), $e_{10}$ $\sim$ \textit{O}($10^0$), $e_{13}$ $\sim$ \textit{O}($10^0$), and $e_{14}$ $\sim$ \textit{O}($10^0$), respectively. To avoid having excessively large terms overshadow other components which could lead to unstable training (see details in \ref{Appendix A}), each loss term is scaled to the same OOM \textit{O}($\mathrm{10^0}$). That is to say, the weight parameters $\lambda_1$ = $10^0$, $\lambda_2$ = $10^{-8}$, $\lambda_3$ = $10^{-8}$, $\lambda_6$ = $10^0$, $\lambda_7$ = $10^0$, $\lambda_{10}$ = $10^0$, $\lambda_{13}$ = $10^0$ and $\lambda_{14}$ = $10^0$, and each loss term is considered to be of equal importance. Note that a large velocity gradient lies in the junction domain of wall and inlet boundary, which would affect the precision of the inlet vertical velocity while training, hence $\lambda_6$ is rewrote as $10^2$ to guarantee the inlet condition and other weights remain unchanged.
In practice, the determination of $\lambda_6$ requires careful consideration. A relevant systematic sensitivity analysis is available in \ref{Appendix A}.
\begin{figure}[htbp]
  \centering
    \includegraphics[scale=0.6]{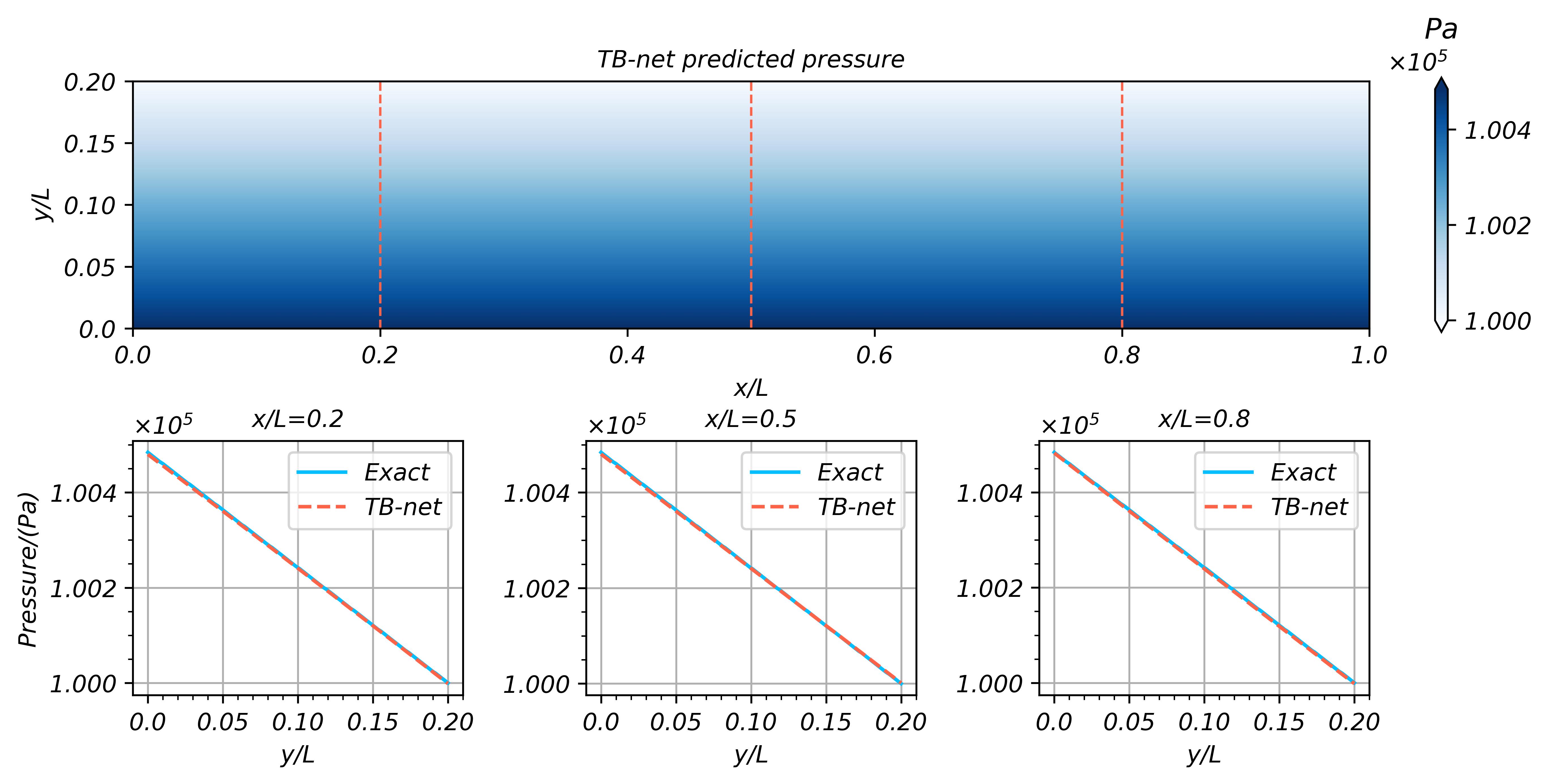}
    \caption{Flow problem with $\dot m$ = 0.1 kg/($\mathrm{m^2}$s): Top: TB-net predicted pressure in the computational domain. Bottom: Comparison of the predicted and exact pressure pertaining to three spatial slices described by the vertical dashed lines in the top panel. The relative $\mathcal{L}_2$ error for the three slices from left to right is 2.147 $\times$ $10^{-5}$, 1.812 $\times$ $10^{-5}$, and 2.195 $\times$ $10^{-5}$, respectively. The relative $\mathcal{L}_2$ error for the entire computational domain is 2.063 $\times$ $10^{-5}$.}\label{fig3}
\end{figure}

In the setup of the example, the training set comprises $N_{Inlet}$ = 400, $N_{Outlet}$ = 400 and $N_{Wall}$ = 100 for both sides to enforce the boundary conditions, and $N_{PDE}$ = 20000 for enforcing the PDEs inside the computational domain. In particular, the collocation points are randomly generated resorting to a space filling Latin hypercube sampling (LHS) strategy. To infer the entire solution $\tilde{u}$, $\tilde{v}$, and $\tilde{p}$ in the computational domain, a 4-layer deep trunk net is devised with 100 neurons per layer, and the branch nets corresponding to the outputs $\tilde{u}$, $\tilde{v}$, and $\tilde{p}$ are 2-layer deep with 50 neurons per layer, respectively. The first hidden layer in trunk net uses a sine activation function to avoid the local minimum trap \cite{wong2022learning}, and the rest layers employ a hyperbolic tangent activation function. Generally speaking, the approximation ability of the NNs should accommodate the complexity of the solution space. In the training process, Adam and L-BFGS optimizers are utilized. Adam optimizer leverages the Jacobian matrix which contains the first-order derivatives of the loss function, while L-BFGS optimizer utilizes the Hessian matrix containing the second-order derivatives of the loss function, hence the L-BFGS converges faster than Adam. However, for stiff and complex scenarios, L-BFGS would be way more easily trapped into local minima. Therefore, in this case, Adam optimizer is firstly used for $1\times10^5$ epochs with learning rate 1\textit{e}-4 to search global minima, and followed by the L-BFGS optimizer until reaching the threshold criterion.
\begin{table*}[pos=!htbp,width=0.7\textwidth]
\caption{The relative $\mathcal{L}_2$ error for the resulting pressure prediction corresponding to different scopes in flow problem with $\dot m$ = 0.5 and 1.0 kg/($\mathrm{m^2}$s), respectively.}\label{tbl3}
\begin{tabular*}{\tblwidth}{@{}LLLLL@{}}
\toprule
 \textbf{Case} & \textit{x} / \textit{L} = 0.2 & \textit{x} / \textit{L} = 0.5 & \textit{x} / \textit{L} = 0.8 & Entire domain \\ 
\midrule
 $\dot m$ = 0.5 kg/($\mathrm{m^2}$s) & 4.134 $\times$ $10^{-5}$ & 1.479 $\times$ $10^{-5}$ & 7.062 $\times$ $10^{-5}$ & 6.618 $\times$ $10^{-5}$ \\
 $\dot m$ = 1.0 kg/($\mathrm{m^2}$s) & 8.218 $\times$ $10^{-5}$ & 4.768 $\times$ $10^{-5}$ & 1.013 $\times$ $10^{-4}$ & 8.372 $\times$ $10^{-5}$ \\
\bottomrule
\end{tabular*}
\end{table*}

In this case, the performance of TB-net PINN for multi-physics solutions with different OOM is highlighted. Taking pressure prediction as an illustration, the relative $\mathcal{L}_2$ error is computed against the test set across the entire computational domain, as well as at three specific spatial slices, i.e., $x/L$ = 0.2, $x/L$ = 0.5 and $x/L$ = 0.8, as shown in Fig. \ref{fig3}. It is evident that the TB-net PINN approach enables effective high-accuracy forecasting.

\textbf{Case B}. In this case, the mass flux $\dot m$ = 0.5 kg/($\mathrm{m^2}$s), and except for the characteristic velocity \textit{V} = 0.5/$\rho$, the other characteristic parameters are consistent with \textit{case} A. In the same manner, the weights corresponding to the loss terms are determined by OOM and importance analysis. Concretely, $\lambda_1$ = $10^0$, $\lambda_2$ = $10^{-10}$, $\lambda_3$ = $10^{-10}$, $\lambda_6$ = $10^2$, $\lambda_7$ = $10^0$, $\lambda_{10}$ = $10^0$, $\lambda_{13}$ = $10^0$ and $\lambda_{14}$ = $10^0$. The configuration of collocation points, NN structure and optimizers is in alignment with \textit{case} A. Analogously, the relative $\mathcal{L}_2$ error for the resulting pressure prediction is assessed both in the whole domain and three typical cross profile, as shown in Table \ref{tbl3}. To our knowledge, the OOM of the errors reached a dramatically small level.

\textbf{Case C}. In this case, the mass flux $\dot m$ is changed to 1.0 kg/($\mathrm{m^2}$s) with the characteristic velocity \textit{V} = 1.0/$\rho$. The corresponding weight parameters are decided as $\lambda_1$ = $10^0$, $\lambda_2$ = $10^{-10}$, $\lambda_3$ = $10^{-10}$, $\lambda_6$ = $10^2$, $\lambda_7$ = $10^0$, $\lambda_{10}$ = $10^0$, $\lambda_{13}$ = $10^0$ and $\lambda_{14}$ = $10^0$. Table \ref{tbl3} illustrates the prediction performance with $\dot m$ = 1.0 kg/($\mathrm{m^2}$s), including both overall and cross profile $\mathcal{L}_2$ analyses.
\begin{figure}[htbp]
  \centering
    \includegraphics[scale=0.55]{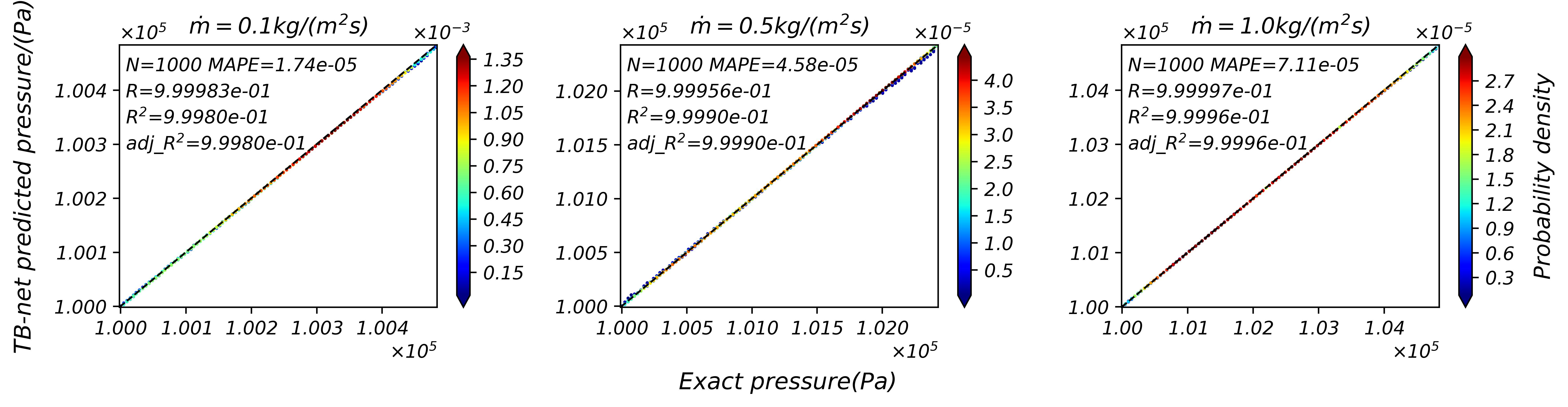}
    \caption{Flow problem: The TB-net predicted and exact pressure distribution and counterpart probability density. Each subgraph has a 1:1 reference line.}\label{fig4}
\end{figure}

It is worth noting that these three cases for the flow problems mainly describe the performance in the whole domain and some typical cross profiles, and for the sake of generality, 1000 spatial points are randomly selected to study the distributions of the prediction and exact solution. As shown in Fig. \ref{fig4}, four extra evaluation metrics are chosen, namely mean absolute percentage error (MAPE), correlation coefficient \textit{R}, determination coefficient $R^2$, and adjusted determination coefficient (adjusted $R^2$). R serves as a statistical quantity to quantify the intensity of the linear relationship existing between two variables, and as it approaches 1, the intensity of the linear relationship increases. $R^2$ is an indicator of how well the model fits the exact solution, and the closer its value to 1, the better the model predicts. The impact of the sample quantity could be balanced out with the implementation of the adjusted $R^2$. In addition to the previously mentioned metrics, kernel density estimation (KDE), a non-parametric estimation technique is also leveraged, to compute the probability density at distribution points, which provides a straightforward visualization of the distribution trends exhibited by both predicted and exact values. From \textit{case} A to C, the data distribution points closely adhere to the 1:1 reference line, indicating robust linearity, and the MAPE reflect the high predictive accuracy of the model.

\begin{figure}[htbp]
  \centering
    \includegraphics[scale=0.55]{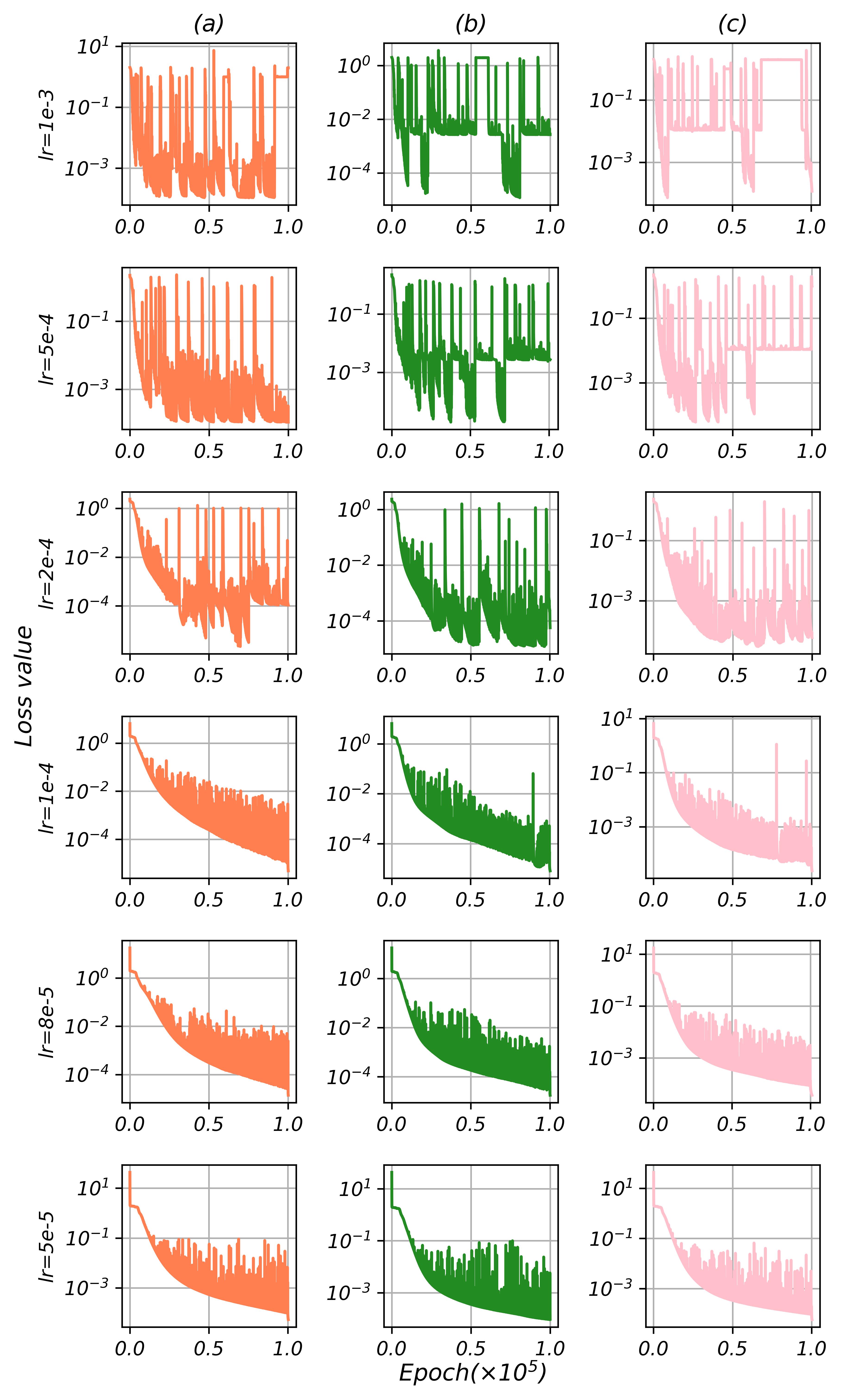}
    \caption{Flow problem loss curve comparison: (a): The loss value curves corresponding to the case 0.1 kg/($\mathrm{m^2}$s) with learning rate (lr) 1\textit{e}-3, 5\textit{e}-4, 2\textit{e}-4, 1\textit{e}-4, 8\textit{e}-5, and 5\textit{e}-5, respectively. (b): The loss value curve with respect to the case $\dot m$ = 0.5 kg/($\mathrm{m^2}$s). (c): The loss value curve with respect to the case $\dot m$ = 1.0 kg/($\mathrm{m^2}$s).}\label{fig5}
\end{figure}

According to the prediction results manifested in the aforementioned cases, the superiority of the OOM and importance analysis is demonstrated in the process of weights assignment. Besides, the complexity of the research problem should be underscored through systematic tuning of the learning rate, and Fig. \ref{fig5} displays the loss curves during the training process with different learning rates. It could be evidently observed that at a larger learning rate, the loss curves of all different cases exhibit unstable oscillation, which reveals that it is not rational for the issue on hand. When the learning rate is smaller, the loss curve shows a relatively smooth descend trend, but it is more likely to cause the training process trapped in a local minimum. In consideration of both stability and effectiveness, learning rate is chosen as 1\textit{e}-4 in all cases involved in this work. It is worth emphasizing that a learning rate of 1\textit{e}-3 proves adequate for the vast majority of classical machine learning scenarios, and its performance in this specific problem implies the complexity of our issue.

\begin{figure}[htbp]
  \centering
    \includegraphics[scale=0.55]{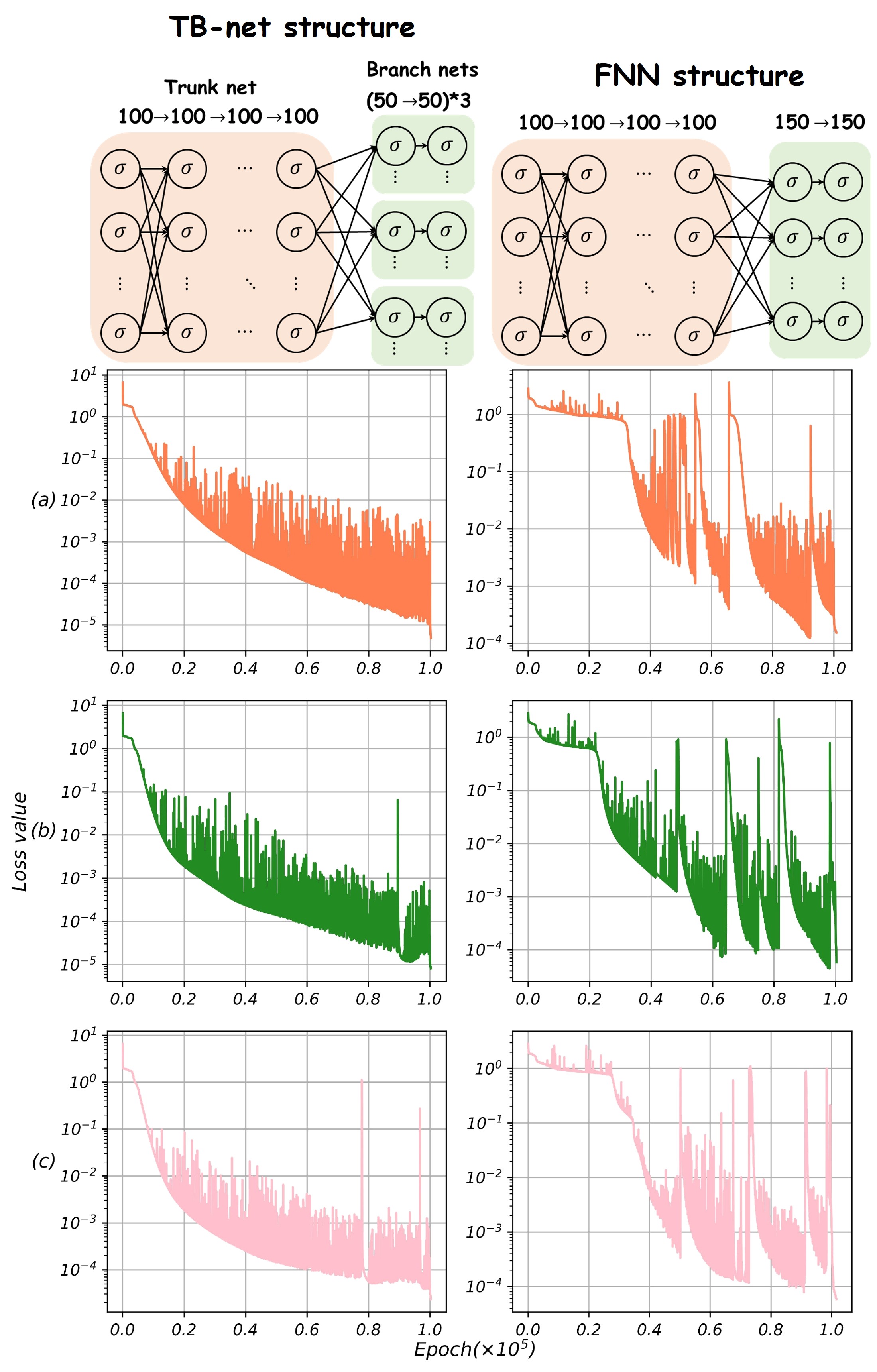}
    \caption{Flow problem loss curve comparison: (a) The loss value curves corresponding to the case $\dot m$ = 0.1 kg/($\mathrm{m^2}$s) with TB-net and FNN, respectively. (b) The loss value curves with respect to the case $\dot m$ = 0.5 kg/($\mathrm{m^2}$s). (c) The loss value curves pertaining to the case $\dot m$ = 1.0 kg/($\mathrm{m^2}$s).}\label{fig6}
\end{figure}

To further demonstrate the capability of TB-net PINN, the training process and predictive accuracy of the PINN with FNN and TB-net are compared, respectively. Note that the structure of FNN here is equivalent to sequentially connecting the trunk part and a vertically integration of three branch parts, hence the quantity of neurons in both models is identical. Fig. \ref{fig6} shows the loss curves of TB-net and FNN under the same training settings, and the training procedure of TB-net is obviously more stable and in-depth compared to FNN in all three cases. Moreover, as FNN has more connections inside compared to TB-net, it possesses more training parameters, which means that the TB-net model consumes less memory and trains faster than FNN. In order to quantify the model predictive accuracy, two metrics are picked, i.e., relative $\mathcal{L}_2$ error and max relative error, and it is evident that the precision of TB-net PINN is significantly superior under both metrics, as shown in Fig. \ref{fig7}. A more detailed quantitative analysis is provided in \ref{Appendix B} for reference.

Additionally, it is notable that the previous discussions have exclusively focused on the pressure prediction, without delving into velocity, and this is owing to the fact that the vertical velocity field involved in this work is virtually uniform, seeming simple at first glance. In fact, attaining high predictive accuracy is challenging due to the disparity in OOM compared with pressure and the presence of steep gradients along the left and right ends. Here, taking the velocity \textit{v} forecasting in the case B as an instance, the relative $\mathcal{L}_2$ error, root mean square error and max relative error are 1.659 $\times$ $10^{-4}$, 8.557 $\times$ $10^{-8}$ and 4.160 $\times$ $10^{-7}$, respectively, and the absolute error could be kept within a small range even at the edges as shown in Fig. \ref{fig8}. While homogeneous porous medium is adopted in this work as an example study, it is clear that our TB-net PINN is capable to conduct high-precision forecasting. Considering that non-uniform porous media are more frequently encountered in real-world applications, our forthcoming work will focus on gradient porosity materials.

\begin{figure}[htbp]
  \centering
    \includegraphics[scale=0.65]{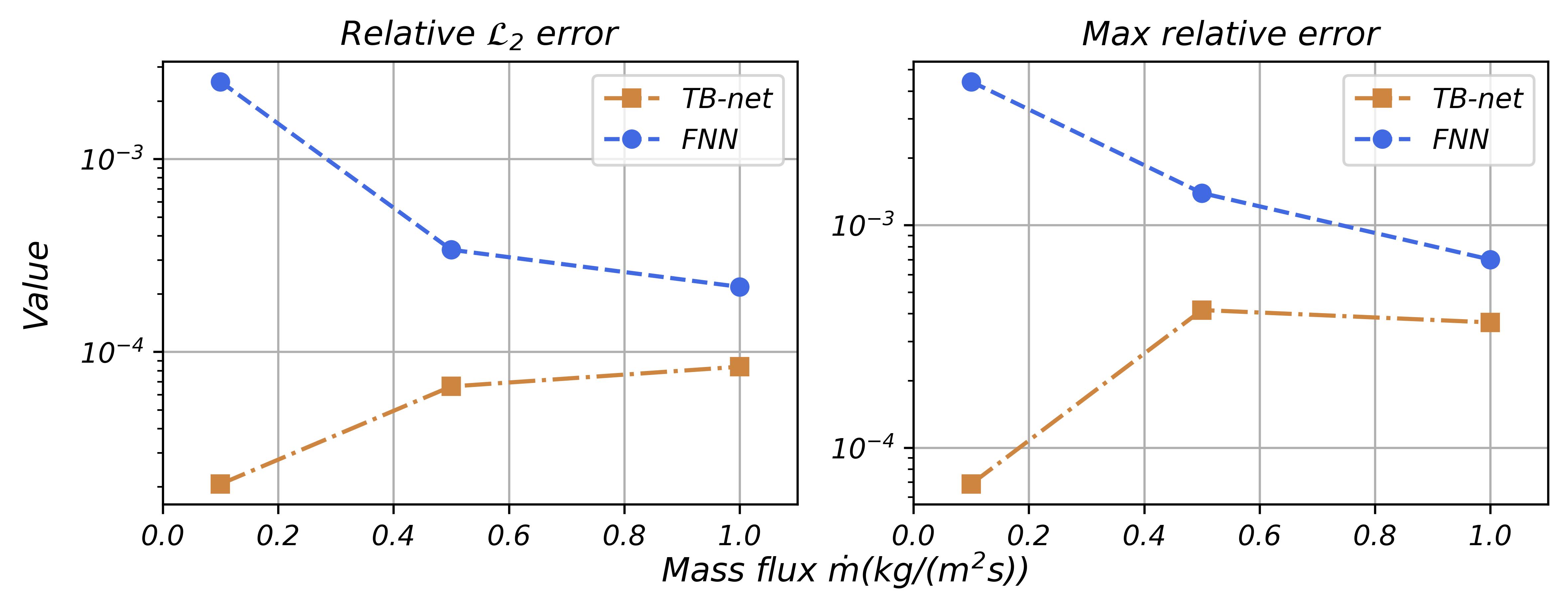}
    \caption{The relative $\mathcal{L}_2$ and max relative error of pressure predicted from TB-net and FNN with respect to the exact pressure.}\label{fig7}
\end{figure}

\begin{figure}[htbp]
  \centering
    \includegraphics[scale=0.5]{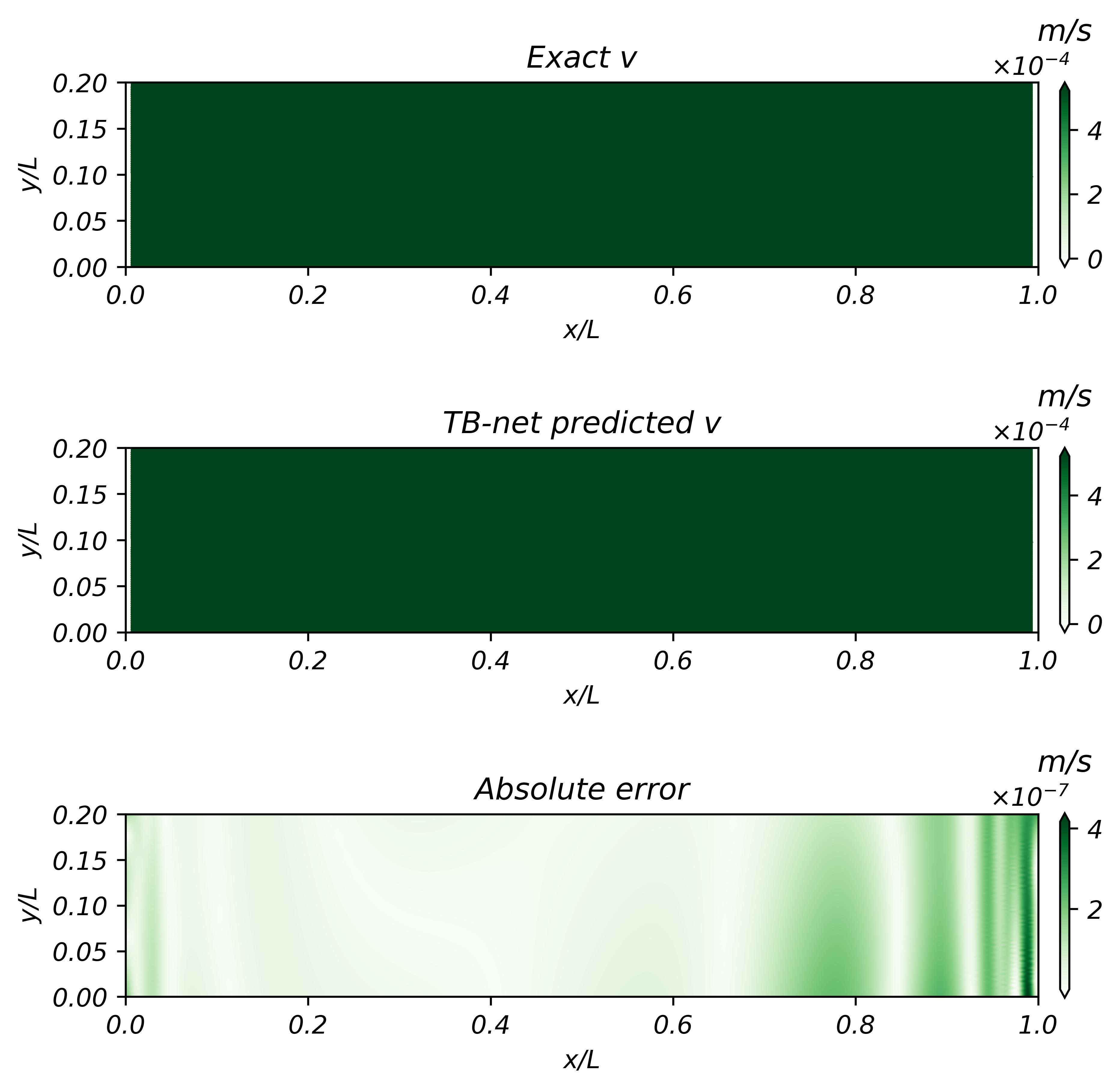}
    \caption{The exact and TB-net predicted vertical velocity \textit{v} and corresponding absolute error for the case $\dot m$=0.5kg/($m^2$s). The relative $\mathcal{L}_2$ error, root mean square error (RMSE) and max absolute error of the computational domain are 1.659 $\times$ $10^{-4}$, 8.557$\times$$10^{-8}$ and 4.160$\times$$10^{-7}$, respectively.}\label{fig8}
\end{figure}
The preceding discussions primarily focused on two-dimensional steady-state flow in porous media. Information on dealing with three-dimensional or even transient flows can be found in \ref{Appendix C}.
\subsubsection{Heat transfer problem}\label{Section 3.1.2}
In this subsection, the focus is temperature field forecasting based on the LTNE model, the solid temperature $T_s$ and fluid temperature $T_f$ are discussed separately. The cases shown below are mainly designed to highlight the flexibility of TB-net structure, with which the problem could be transformed from flow pattern to heat transfer pattern smoothly.
Compared to flow problems, heat transfer problems require extra consideration of energy equations and thermal boundary conditions, including Eqs.(\ref{Eq.3d}-\ref{Eq.3e}), (\ref{Eq.5c}-\ref{Eq.5d}), (\ref{Eq.5f}-\ref{Eq.5g}) and (\ref{Eq.5j}-\ref{Eq.5k}). That is to say, the loss function of heat transfer problems involves loss terms $e_4$ $\sim$ $e_5$, $e_8$ $\sim$ $e_9$, $e_{11}$ $\sim$ $e_{12}$, and $e_{15}$ $\sim$ $e_{16}$, additionally. Note that here the dynamic viscosity and Prandtl number are postulated as value at the reference temperature 300 K, hence the flow field would remain unchanged regardless of whether heat transfer effects are considered. Consequently, once the flow problem is solved, the heat transfer calculations could be proceeded based on its results. For example, three cases are conducted in this part, i.e., \textit{case} D, E and F with respect to 5.0 $\times$ $10^4$, 1.0 $\times$ $10^5$ and 1.5 $\times$ $10^5$ W/$\mathrm{m^2}$, respectively, and the mass flux remains the same as $\dot m$ = 0.5 kg/($\mathrm{m^2}$s). There is no need to train the entire flow and heat transfer process from scratch. Instead, the pre-trained model from \textit{case} B could be leveraged, freezing the parameters corresponding to the trunk net and branch nets with respect to $\widetilde{u}$, $\widetilde{v}$, and $\widetilde{p}$, and devote training efforts solely to the branch nets pertaining to the $\widetilde{h_k}$ and $\widetilde{T_s}$. The loss function would exclusively employ the extra loss terms mentioned earlier, excluding the loss terms related to flow issues described in Section \ref{Section 3.1.1}. It is worth noting that if all loss terms are trained from scratch rather than adopting this step-wise strategy, the training process would ultimately fail even within the TB-net architecture because of the complexity and multitude of loss terms involved.
A detailed comparison between the step-wise sequential solution mode and the coupled solution mode is provided in \ref{Appendix D}.

\textbf{Case D}. In this case, the heat flux $\dot q$ applied to the outlet boundary is set as 5.0 $\times$ $10^4$ W/$\mathrm{m^2}$, and the temperature of the injected liquid is 300 K, hence the characteristic temperature \textit{T} and kinetic enthalpy \textit{T}$C_P$ is 300 K and 1.2651$\times$$10^6$ J/kg, respectively. The other characteristic parameters \textit{V}, \textit{P} and \textit{L} are consistent with those in \textit{case} B. The OOM of the loss terms is $e_4$ $\sim$ \textit{O}($10^{11}$), $e_5$ $\sim$ \textit{O}($10^{11}$), $e_8$$\sim$ \textit{O}($10^3$), $e_9$ $\sim$ \textit{O}($10^3$), $e_{11}$$\sim$ \textit{O}($10^4$), $e_{12}$ $\sim$ \textit{O}($10^0$), $e_{15}$ $\sim$ \textit{O}($10^0$) and $e_{16}$ $\sim$ \textit{O}($10^0$). The loss terms are firstly normalized to OOM of \textit{O}($10^0$), which means $\lambda_4 = 10^{-22}$, $\lambda_5 = 10^{-22}$, $\lambda_8 = 10^{-6}$, $\lambda_9 = 10^{-6}$, $\lambda_{11} = 10^{-8}$, and $\lambda_{12} = \lambda_{15} = \lambda_{16} = 10^0$. Note that the accuracy of the prediction on the inlet and outlet boundaries is paramount to the overall precision of the computational domain, hence $\lambda_8$, $\lambda_9$, $\lambda_{11}$ and $\lambda_{12}$ are adjusted as $10^{-4}$, $10^{-4}$, $10^{-7}$, and $10^1$, respectively.
\begin{figure}[htbp]
  \centering
    \includegraphics[scale=0.57]{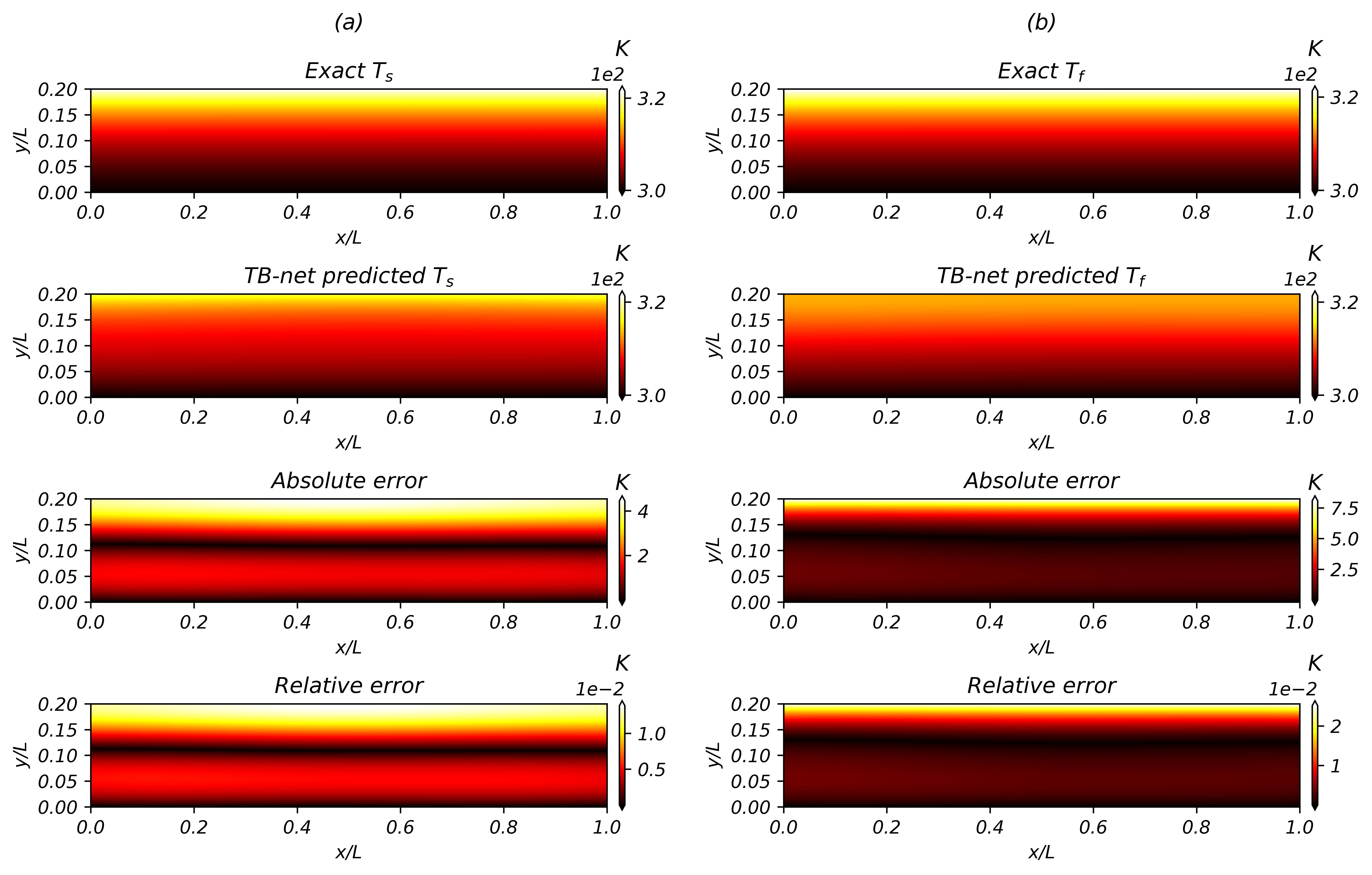}
    \caption{Heat transfer problem with $\dot q$ = 5.0 $\times$ $10^4$ W/$\mathrm{m^2}$: (a) The exact and TB-net predicted solid temperature $T_s$ and corresponding absolute and relative error. (b) The exact and TB-net predicted fluid temperature $T_f$ and corresponding absolute and relative error. The relative $\mathcal{L}_2$ error for the $T_s$ and $T_f$ is 6.872 $\times$ $10^{-3}$ and 7.742 $\times$ $10^{-3}$, respectively, and the counterpart max relative error is 1.386 $\times$ $10^{-2}$ and 2.512 $\times$ $10^{-2}$, respectively.}\label{fig9}
\end{figure}

The branch nets corresponding to outputs $\widetilde{h_k}$ and $\widetilde{T_s}$ are 4-layer deep with 100 neurons per layer, respectively, and meanwhile, the activation functions of their first hidden layers are both set as sine to avoid local minima. It is noteworthy that the branch nets dedicated to $\widetilde{h_k}$ and $\widetilde{T_s}$ are more extensive than that for $\widetilde{u}$, $\widetilde{v}$, and $\widetilde{p}$ previously discussed. This stems from the more intricate loss function pertaining to the energy equation, requiring more robust networks to navigate the larger search space. As the remaining parts of the TB-net architecture employ the pre-trained parameters from \textit{case} B, parameter optimization is solely necessary for the two newly defined branch nets. Additionally, the quantity, generation method of collocation points necessary for training, as well as the parameters of optimizers, remain consistent with the configurations specified for the flow problems.

Fig. \ref{fig9} summarizes the results of this case. It is evident that irrespective of whether the prediction of $T_f$ or Ts, the computational domain exhibits a relative $\mathcal{L}_2$ error of \textit{O}($10^{-3}$) and a max relative error of \textit{O}($10^{-2}$). Therefore, optimizing the energy equation loss with the aid of pre-trained flow model serves as an efficient means to discern complex heat transfer tendencies. It is particularly important to mention that the FNN architecture would directly fail in the heat transfer problems due to the myriad loss terms no matter pre-trained models are utilized or not, which reveals the flexibility and robustness of TB-net.

\textbf{Case E.} In this case, the heat flux boundary condition $\dot q$ is altered as 1.0 $\times$ $10^5$ W/$\mathrm{m^2}$, and the rest of the boundary conditions and characteristic parameters are kept identical to those in \textit{case} D. Owing to the change of thermal boundary condition, $\lambda_{11}$ needs to be modified according to the OOM analysis, ultimately yielding $\lambda_4$ = $10^{-22}$, $\lambda_5$ = $10^{-22}$, $\lambda_8$ = $10^{-6}$, $\lambda_9$ = $10^{-6}$, $\lambda_{11}$ = $10^{-10}$, and $\lambda_{12}$ = $\lambda_{15}$ = $\lambda_{16}$ = $10^0$. Similarly, a full-field error analysis of the temperature prediction results is conducted as shown in Table \ref{tbl4}. 

\begin{table*}[pos=!htbp,width=0.7\textwidth]
\caption{The relative $\mathcal{L}_2$ error and max relative error for the $T_s$ and $T_f$ in heat transfer problem with 1.0 $\times 10^5$ and 1.5 $\times 10^5$ W/$\mathrm{m^2}$, respectively.}\label{tbl4}
\begin{tabular*}{\tblwidth}{@{}LLLLL@{}}
\toprule
   & \multicolumn{2}{c}{$T_s$ error} & \multicolumn{2}{c}{$T_f$ error} \\ 
\midrule
 \textbf{Case} & relative $\mathcal{L}_2$ & max relative & relative $\mathcal{L}_2$ & max relative \\
\midrule
  $\dot q$ = 1.0 $\times 10^5$ W/$\mathrm{m^2}$ & 9.033 $\times 10^{-3}$ & 1.357 $\times 10^{-2}$ & 1.115 $\times 10^{-2}$ & 2.929 $\times 10^{-2}$ \\
   $\dot q$ = 1.5 $\times 10^5$ W/$\mathrm{m^2}$ & 1.103 $\times 10^{-2}$ & 1.586 $\times 10^{-2}$ & 1.249 $\times 10^{-2}$ & 3.450 $\times 10^{-2}$ \\
\bottomrule
\end{tabular*}
\end{table*}
\begin{figure}[htbp]
  \centering
    \includegraphics[scale=0.5]{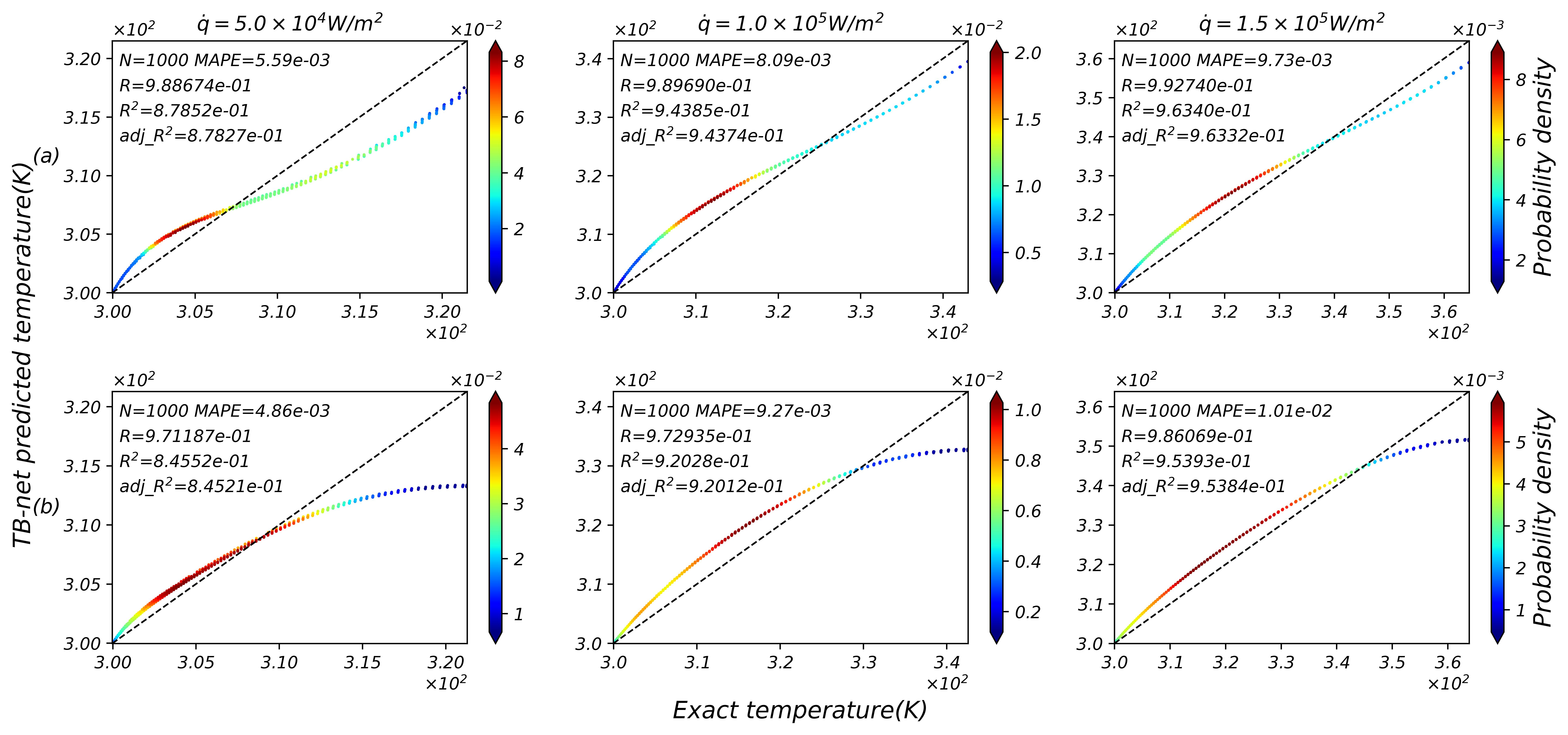}
    \caption{Heat transfer problem: (a) The TB-net predicted and exact $T_s$ distribution and counterpart probability density. (b) The TB-net predicted and exact $T_f$ distribution and counterpart probability density. Each subgraph has a 1:1 reference line.}\label{fig10}
\end{figure}

\textbf{Case F.} In this case, the heat flux on the outlet boundary $\dot q$ = 1.5 $\times$ $10^5$ W/$\mathrm{m^2}$, and based on the OOM analysis, the weight parameters are set the same as in \textit{case} E. Table \ref{tbl4} displays the relevant quantified error analyses.
It is worth highlighting that although the relative $\mathcal{L}_2$ error and max relative error manifested in \textit{case} D, E and F are deemed tolerable within the context of most machine learning tasks, Fig. \ref{fig9} reveals relatively larger local inaccuracies, for example, at the outlet boundary. Consequently, to gain a more intuitive assessment of the effectiveness of the model prediction, a KDE analysis is proceeded.
As shown in Fig. \ref{fig10}, 1000 randomly spatial points are selected in the computational domain, and corresponding MAPE, \textit{R}, $R^2$, and adjusted $R^2$ are calculated to evaluate the linearity and accuracy. The resulting data point distributions illustrate that the predicted and exact values exhibit the greatest discrepancy from the reference line at the outlet boundary, which is primarily attributed to the complex coupling effect of the flow and heat transfer conditions on the outlet boundary. Such issues could be tackled with standard measures like encrypting local collocation points. Fortunately, with the advancements in experimental techniques, especially the ITIS for the surface temperature measurements, the outer surface temperature could be leveraged to address the inverse problems.

\subsection{Inverse problems}\label{Section 3.2}
In contrast to the forward problem, the inverse problem encounters a scarcity of sufficient prerequisites for direct resolution, and here the absence of heat flux condition on the outlet boundary is taken as an illustration. In this situation, extra labeled data is required in conjunction with physical laws to yield solutions. In terms of implementation, PINNs-like methods could solve inverse problems as easily as forward problems with slight code changes. Furthermore, with the extra constraints of data, the convergence of the training process is significantly improved, resulting in a substantial increase in prediction accuracy. Nevertheless, traditional numerical methods face prominent challenges in achieving this, often at a prohibitive cost or as an impractical task.
\begin{figure}[htbp]
  \centering
    \includegraphics[scale=0.4]{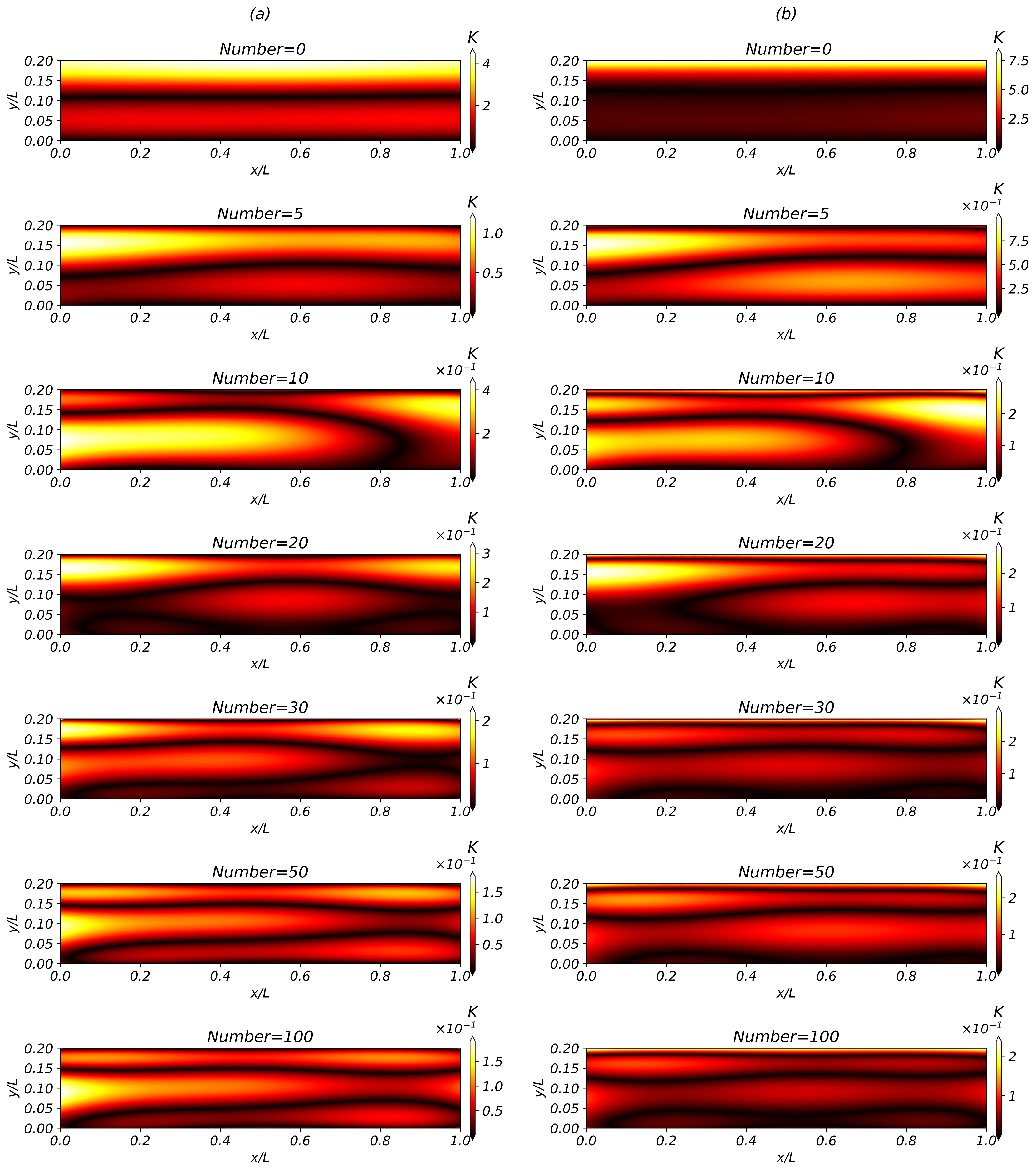}
    \caption{Inverse heat transfer problem with $\dot q$ = 5.0 $\times$ $10^4$ W/$\mathrm{m^2}$: (a) The absolute error of $T_s$ with respect to different number of extra labeled data. (b) The absolute error of $T_f$ with respect to different number of extra labeled data.}\label{fig11}
\end{figure}

Here, three inverse problem investigations pertaining to \textit{case} D, E and F are focused on. Distinct from forward problems, the loss function $\mathcal{L}(\sigma)$ incorporates data loss $\mathcal{L}_{Data}$, with components comprising the loss terms $e_{17}$ and $e_{18}$. Generally, data utilized in solving inverse problems is optimally derived from experimental measurements, such as ITIS. Nonetheless, simulated data is primarily employed to validate the practicability of the algorithm, aiming to provide guidance for engineering applications. A series of amounts of outlet boundary data points are randomly selected, i.e., 0, 5, 10, 20, 30, 50, and 100, to explore their impact on the accuracy of inverse problem solution. It is noteworthy that the provided additional data points is considered to be of high fidelity, making the loss terms $e_{17}$ and $e_{18}$ more crucial, hence the corresponding weights are designated as $\lambda_{17} = 10^2$ and $\lambda_{18} = 10^2$. Meanwhile, the lack of heat flux conditions results in the exclusion of $e_{11}$ and $e_{12}$ in $\mathcal{L}(\sigma)$. The remaining parameter settings mirror exactly those specified in the corresponding forward problems. Note that the restrictive effect of $\mathcal{L}_{Data}$ exerts on the search space essentially accelerate convergence, thus reducing the overall computational cost.

Fig. \ref{fig11} presents the absolute error of $T_s$ and $T_f$ with respect to different number of extra labeled data points corresponding to the inverse investigation pertaining to \textit{case} D, and the overall trend is analogous in relevant inverse problems with respect to \textit{case} E and F. It is apparent that the magnitude of the absolute error undergoes a significant decline with the augmentation of the specified labeled data points. For more detailed analyses, the relative $\mathcal{L}_2$ errors and max relative errors of $T_s$ and $T_f$ are compared, respectively.

\begin{figure}[htbp]
  \centering
    \includegraphics[scale=0.6]{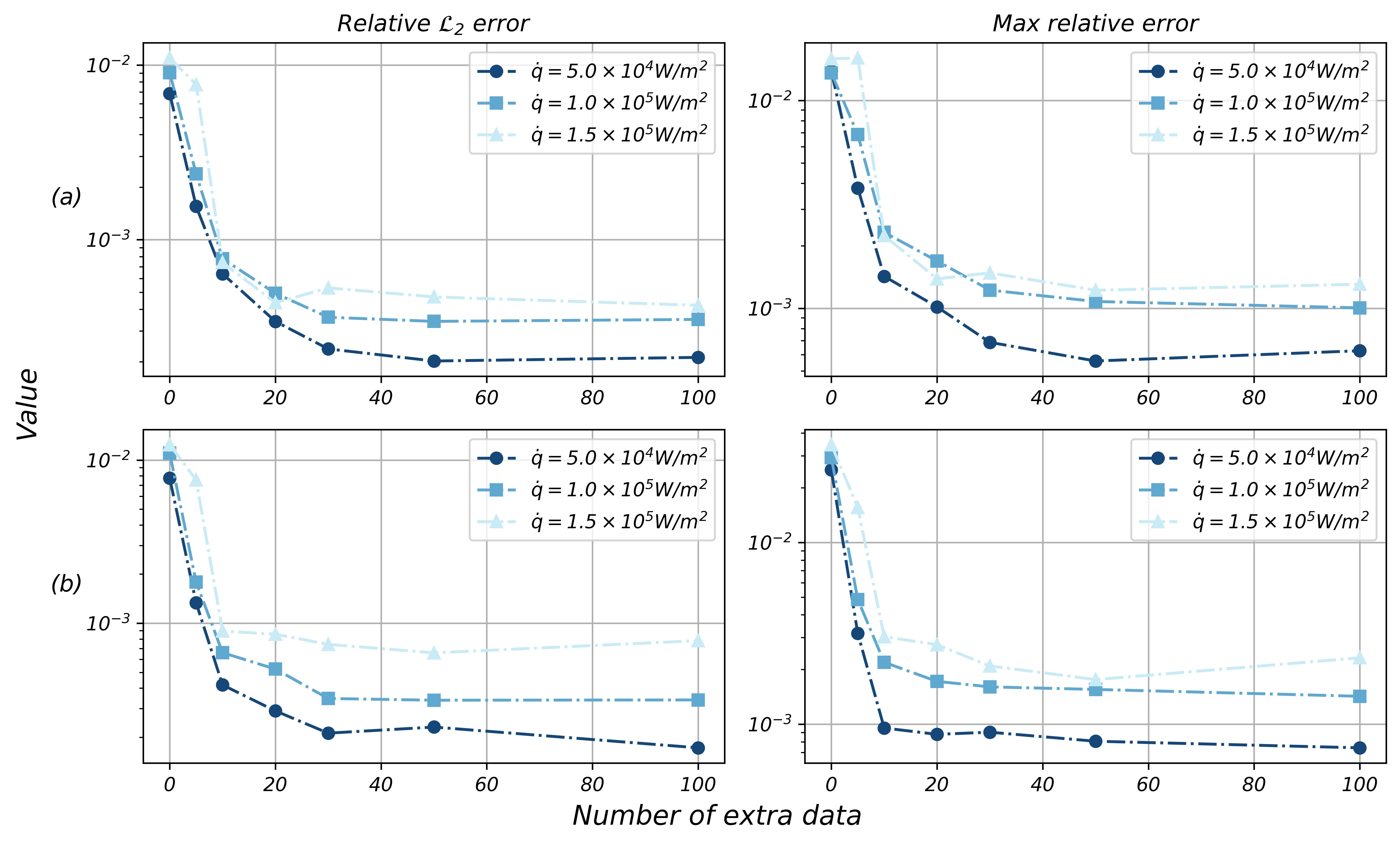}
    \caption{Inverse heat transfer problem: (a) The relative $\mathcal{L}_2$ error and max relative error of $T_s$. (b) The relative $\mathcal{L}_2$ error and max relative error of $T_f$.}\label{fig12}
\end{figure}

\begin{figure}[htbp]
  \centering
    \includegraphics[scale=0.3]{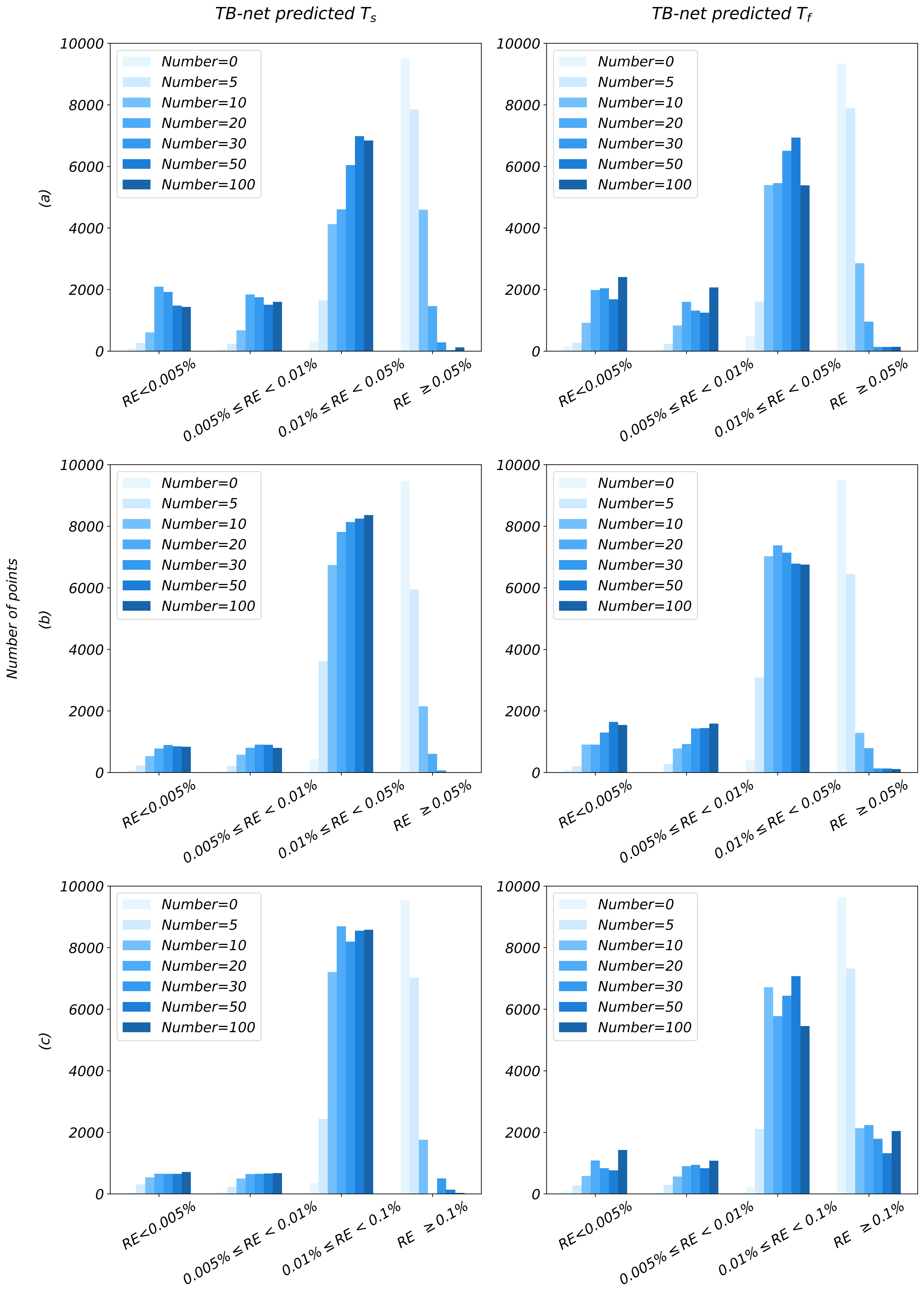}
    \caption{The number of points corresponding to RE scope of TB-net predicted $T_s$ and $T_f$ : (a) Inverse heat transfer problem with $\dot q$ = 5.0 $\times$ $10^4$ W/$\mathrm{m^2}$. (b) Inverse heat transfer problem with $\dot q$ = 1.0 $\times$ $10^5$ W/$\mathrm{m^2}$ (c) Inverse heat transfer problem with $\dot q$ = 1.5 $\times$ $10^5$ W/$\mathrm{m^2}$.}\label{fig13}
\end{figure}

As shown in Fig. \ref{fig12}, when number of labeled data equals 0 (the forward problem without labeled data), the relative $\mathcal{L}$ error and max relative error peak, and the errors diminish considerably with minor increments in the number. Subsequently, as the number ranges from 20 to 100, the errors stabilize at a low level. A statistical analysis is also conducted on the number of points within different intervals, utilizing relative error (RE) as the basis. Fig. \ref{fig13} provide straightforward depictions of the changing pattern in the number of points distributed across different scopes, contingent upon the varying amount of labeled given data points. Without exception, when the given additional data points are few, the number of points falling within the scope of maximum RE decreases significantly, while the number within the range of minimum RE increases substantially, and there is a notable improvement in overall prediction accuracy. As the amount of labeled data points reaches 20 or more, the distribution of data points within each error range tends to stabilize.

\begin{figure}[htbp]
  \centering
    \includegraphics[scale=0.5]{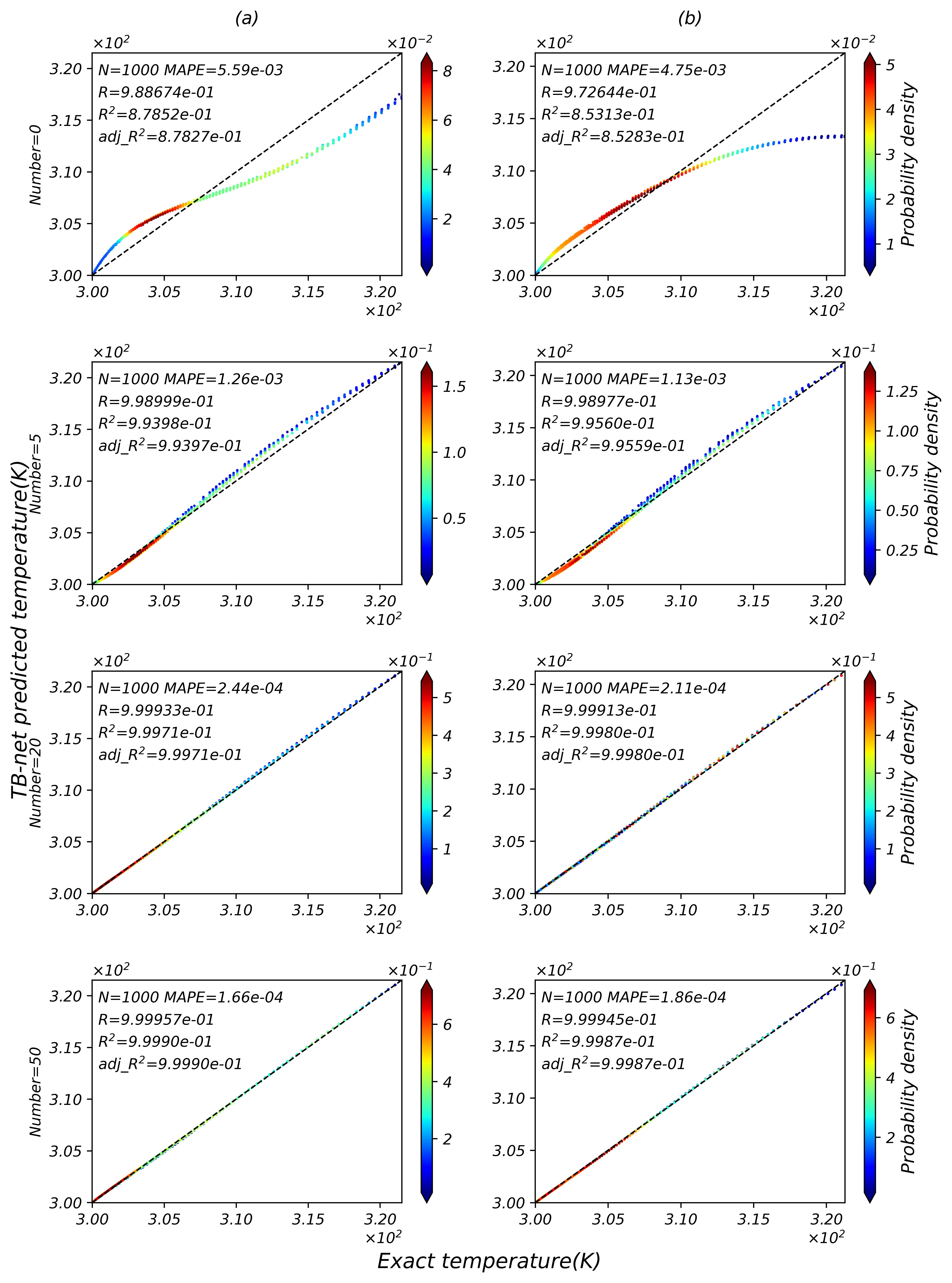}
    \caption{Inverse heat transfer problem with $\dot q$ = 5.0 $\times$ $10^4$ W/$\mathrm{m^2}$: (a) The TB-net predicted and exact $T_s$ distribution and counterpart probability density. (b) The TB-net predicted and exact $T_f$ distribution and counterpart probability density.}\label{fig14}
\end{figure}

In addition, KDE analyses are performed on cases with number of labeled data points equals 0, 5, 20, 50. As shown in Fig. \ref{fig14}, in the case where there are zero additional data points, i.e., the forward problem, notable deviations occur between the predicted and exact values at the outlet boundary, and regions characterized by higher probability density exhibit a clear separation from the reference line. Compared to having zero labeled points, having 5 results in a remarkable enhancement in precision at the boundaries, with the distribution of points aligning more closely with the reference line. Moreover, the regions of high probability density nearly coincide with the reference line, and metrics assessing correlation and linearity, i.e., MAPE, \textit{R}, $R^2$ and adjusted $R^2$, display a prominent improvement. Upon examining the scenario where the amount of the labeled data points equals 20, it is noticeable that the vast majority of the data points are situated directly on the reference line. Concurrently, the statistical metrics, which are already at a commendable level for 5 labeled data points case, have even realized a further boost. Following this, the data distribution and statistical indicators for the case where amount of extra labeled data points equals 50 are delved into, observing that their overall performance is similar with that of amount equals 20. The trends of KDE analysis in the inverse heat problems with respect to $\dot q$ = 1.0 $\times$ $10^5$ and $\dot q$ = 1.5 $\times$ $10^5$ W/$\mathrm{m^2}$ are similar to those in Fig. \ref{fig14}, and the corresponding statistical metrics are summarized in Table \ref{tbl5}.
According to the earlier quantitative analyses on how varying the amount of extra labeled points affect prediction effectiveness, substantial enhancements in prediction accuracy can be achieved with few additional points could be concluded. Nevertheless, the accuracy would not escalate indefinitely with the augmentation of the labeled data points, it stabilizes when the number of labeled points reaches 20 or greater. Therefore, for practical engineering purposes, vastly improved predictions could be made based on minimal known data. Notably, even when the need for extremely high prediction accuracy arises, incorporating only 20 additional data points can yield similar improvements as using 100, thereby significantly reducing experimental expenses and burdens.
However, the data obtained during experimental processes inevitably contain random errors. In this context, evaluations under noisy conditions are conducted in \ref{Appendix E}.

\begin{table*}[pos=!htbp,width=0.7\textwidth]
\caption{The statistical metrics for different number of labeled points in inverse heat transfer problems with $\dot q$ = 1.0 $\times$ $10^5$ W/$\mathrm{m^2}$ and $\dot q$ = 1.5 $\times$ $10^5$ W/$\mathrm{m^2}$, respectively.}\label{tbl5}
\begin{tabular*}{\tblwidth}{@{}LLLLL@{}}
\toprule
   & \multicolumn{4}{c}{$\dot q$ = 1.0 $\times$ $10^5$ W/$\mathrm{m^2}$ for $T_s$} \\ 
\midrule
 {Metrics} & number = 0 & number = 5 & number = 20 & number = 50 \\
\midrule
  MAPE & 8.20 $\times$ $10^{-3}$ & 1.71 $\times$ $10^{-3}$ & 3.56 $\times$ $10^{-4}$ & 2.86 $\times$ $10^{-4}$ \\
   \textit{R} & 9.9023 $\times$ $10^{-1}$ & 9.9910 $\times$ $10^{-1}$ & 9.9996 $\times$ $10^{-1}$ & 9.9997 $\times$ $10^{-1}$ \\
   $R^2$ & 9.4611 $\times$ $10^{-1}$ & 9.9646 $\times$ $10^{-1}$ & 9.9985 $\times$ $10^{-1}$ & 9.9993 $\times$ $10^{-1}$ \\
   Adjusted $R^2$ & 9.4600 $\times$ $10^{-1}$ & 9.9645 $\times$ $10^{-1}$ &9.9985 $\times$ $10^{-1}$	& 9.9993 $\times$ $10^{-1}$ \\
\midrule
& \multicolumn{4}{c}{$\dot q$ = 1.0 $\times$ $10^5$ W/$\mathrm{m^2}$ for $T_f$} \\
\midrule
 {Metrics} & number = 0 & number = 5 & number = 20 & number = 50 \\
 \midrule
MAPE & 9.53 $\times$ $10^{-3}$ & 1.44 $\times$ $10^{-3}$ & 3.65 $\times$ $10^{-4}$ & 2.56 $\times$ $10^{-4}$ \\
\textit{R} & 9.7182 $\times$ $10^{-1}$ & 9.9944 $\times$ $10^{-1}$ & 9.9994 $\times$ $10^{-1}$ & 9.9998 $\times$ $10^{-1}$ \\
$R^2$ & 9.1286 $\times$ $10^{-1}$ & 9.9791 $\times$ $10^{-1}$ & 9.9982 $\times$ $10^{-1}$ & 9.9992 $\times$ $10^{-1}$ \\
Adjusted $R^2$ & 9.1268 $\times$ $10^{-1}$ & 9.9790 $\times$ $10^{-1}$ & 9.9982 $\times$ $10^{-1}$ & 9.9992 $\times$ $10^{-1}$ \\
\midrule
& \multicolumn{4}{c}{$\dot q$ = 1.5 $\times$ $10^5$ W/$\mathrm{m^2}$ for $T_s$} \\
\midrule
 {Metrics} & number = 0 & number = 5 & number = 20 & number = 50 \\
 \midrule
MAPE & 9.99 $\times$ $10^{-3}$ & 5.75 $\times$ $10^{-3}$ & 3.72 $\times$ $10^{-4}$ & 3.93 $\times$ $10^{-4}$ \\
\textit{R} & 9.9238 $\times$ $10^{-1}$ & 9.9773 $\times$ $10^{-1}$ & 9.9997 $\times$ $10^{-1}$ & 9.9997 $\times$ $10^{-1}$ \\
$R^2$ & 9.6244 $\times$ $10^{-1}$ & 9.8146 $\times$ $10^{-1}$ & 9.9994 $\times$ $10^{-1}$ & 9.9993 $\times$ $10^{-1}$ \\
Adjusted $R^2$ & 9.6236 $\times$ $10^{-1}$ & 9.8142 $\times$ $10^{-1}$ & 9.9994 $\times$ $10^{-1}$ & 9.9993 $\times$ $10^{-1}$ \\
\midrule
& \multicolumn{4}{c}{$\dot q$ = 1.5 $\times$ $10^5$ W/$\mathrm{m^2}$ for $T_f$} \\
\midrule
 {Metrics} & number = 0 & number = 5 & number = 20 & number = 50 \\
\midrule
MAPE & 1.06 $\times$ $10^{-2}$ & 5.55 $\times$ $10^{-3}$ & 5.69 $\times$ $10^{-4}$ & 4.69 $\times$ $10^{-4}$ \\
\textit{R} & 9.8493 $\times$ $10^{-1}$ & 9.9748 $\times$ $10^{-1}$ & 9.9991 $\times$ $10^{-1}$ & 9.9995 $\times$ $10^{-1}$ \\
$R^2$ & 9.4859 $\times$ $10^{-1}$ & 9.8206 $\times$ $10^{-1}$ & 9.9978 $\times$ $10^{-1}$ & 9.9986 $\times$ $10^{-1}$ \\
Adjusted $R^2$ & 9.4849 $\times$ $10^{-1}$ & 9.8202 $\times$ $10^{-1}$ & 9.9978 $\times$ $10^{-1}$ & 9.9986 $\times$ $10^{-1}$ \\
\bottomrule
\end{tabular*}
\end{table*}

\subsection{Transfer learning}\label{Section 3.3}
Previous discussions primarily revolved around scenarios with fixed porosity, standard atmosphere environment, and fixed heat flux conditions. Once these conditions change, the previously trained model loses its applicability, and initiating a fresh training process from scratch would lead to a significant waste of computational resources. Transfer learning, as a breakthrough in machine learning community, provides a viable path to address it. To be specific, when a model is trained, it learns some high-level abstractions, which could be applied to analogous tasks. When faced with such tasks, they could be solved efficiently by fine-tuning partial parameters of the model. In the following, the issues pertaining to different porosities, external pressures, and heat flux conditions are tackled from the perspective of transfer learning.
\subsubsection{Effect of porosity on the flow field}\label{Section 3.3.1}
In this work, sintered particle porous media is employed, and commonly its porosity ranges from 0.2 to 0.5 \cite{ternero2021influence}. The flow problem with a porosity $\varepsilon$ = 0.3 has been discussed previously, and here attention is shifted to flow problems characterized by $\varepsilon$ = 0.4 and $\varepsilon$ = 0.5. In particular, the porosity is altered in \textit{case} B, while ensuring the boundary conditions, including mass flux and outlet pressure, unvaried. At this point, the OOM of all loss terms stays the same, hence the corresponding weight parameters are kept consistent with those in \textit{case} B. It bears mentioning that the training mode comprises two types, i.e., from scratch and transfer learning. In the context of training from scratch, the TB-net structure remains identical to that utilized in \textit{case} B, and the training starts from a randomly initialized state. In contrast, when adopting the transfer learning pattern, the initial parameters of the network are those trained in \textit{case} B. On this basis, only the parameters of the three branch nets undergo further training. This is due to the fact that modifying the porosity is fundamentally a similar task, and the model trained in case B has already learned high-dimensional features. As a result, by merely fine-tuning the model, a swift resolution to the new problem can be achieved. 

\begin{figure}[htbp]
  \centering
    \includegraphics[scale=0.5]{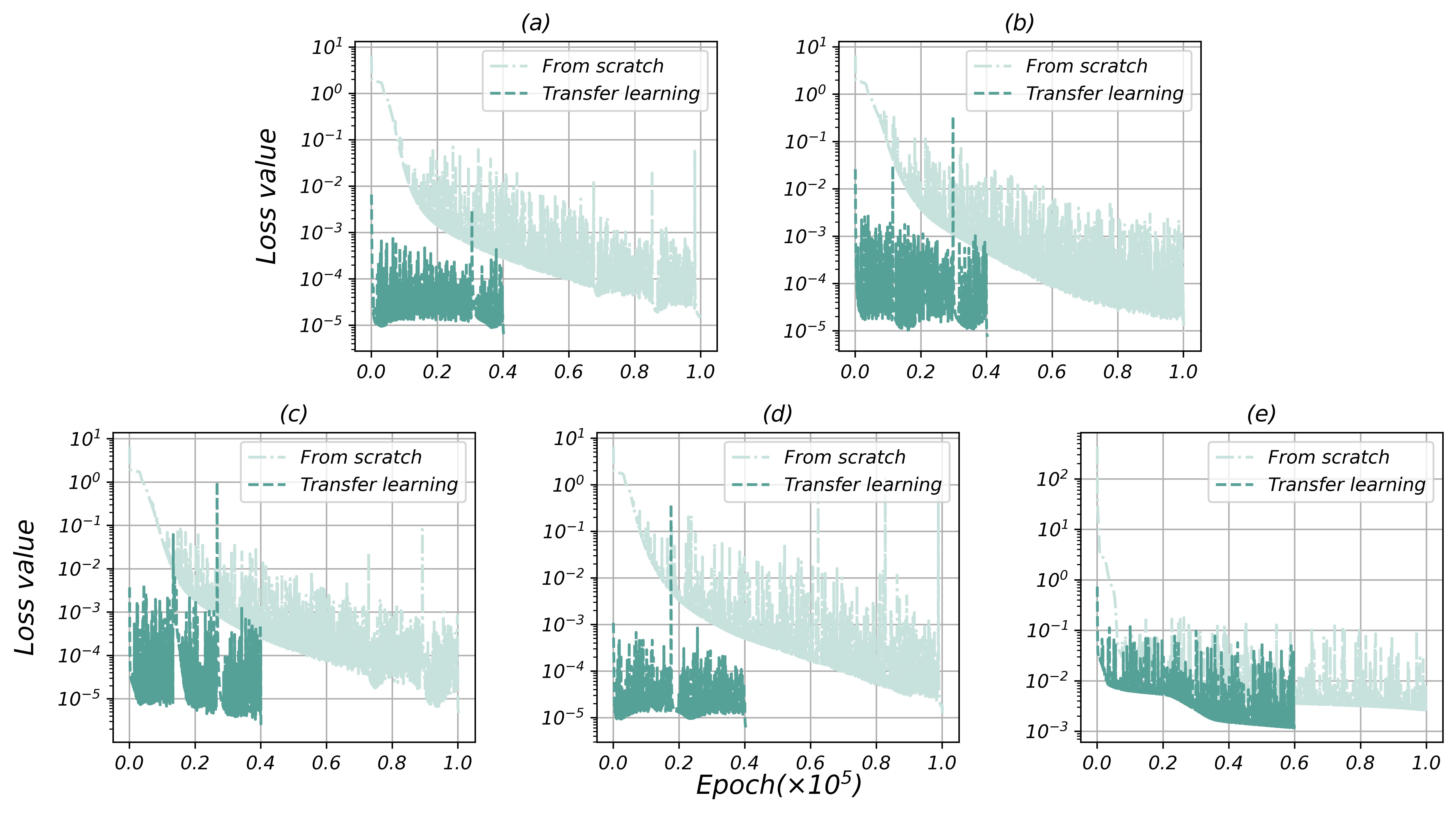}
    \caption{The loss value curves corresponding to TB-net PINN training from scratch and transfer learning: (a) Flow problem with $\varepsilon$ = 0.4. (b) Flow problem with $\varepsilon$ = 0.5. (c) Flow problem with $\dot m$ = 0.1 and $P$ = 5.0 $\times$ $10^4$. (d) Flow problem with $\dot m$ = 0.5 and $P$ = 5.0 $\times$ $10^4$. (e) Inverse heat transfer problem with linear $\dot q$.}\label{fig15}
\end{figure}

\begin{figure}[htbp]
  \centering
    \includegraphics[scale=0.5]{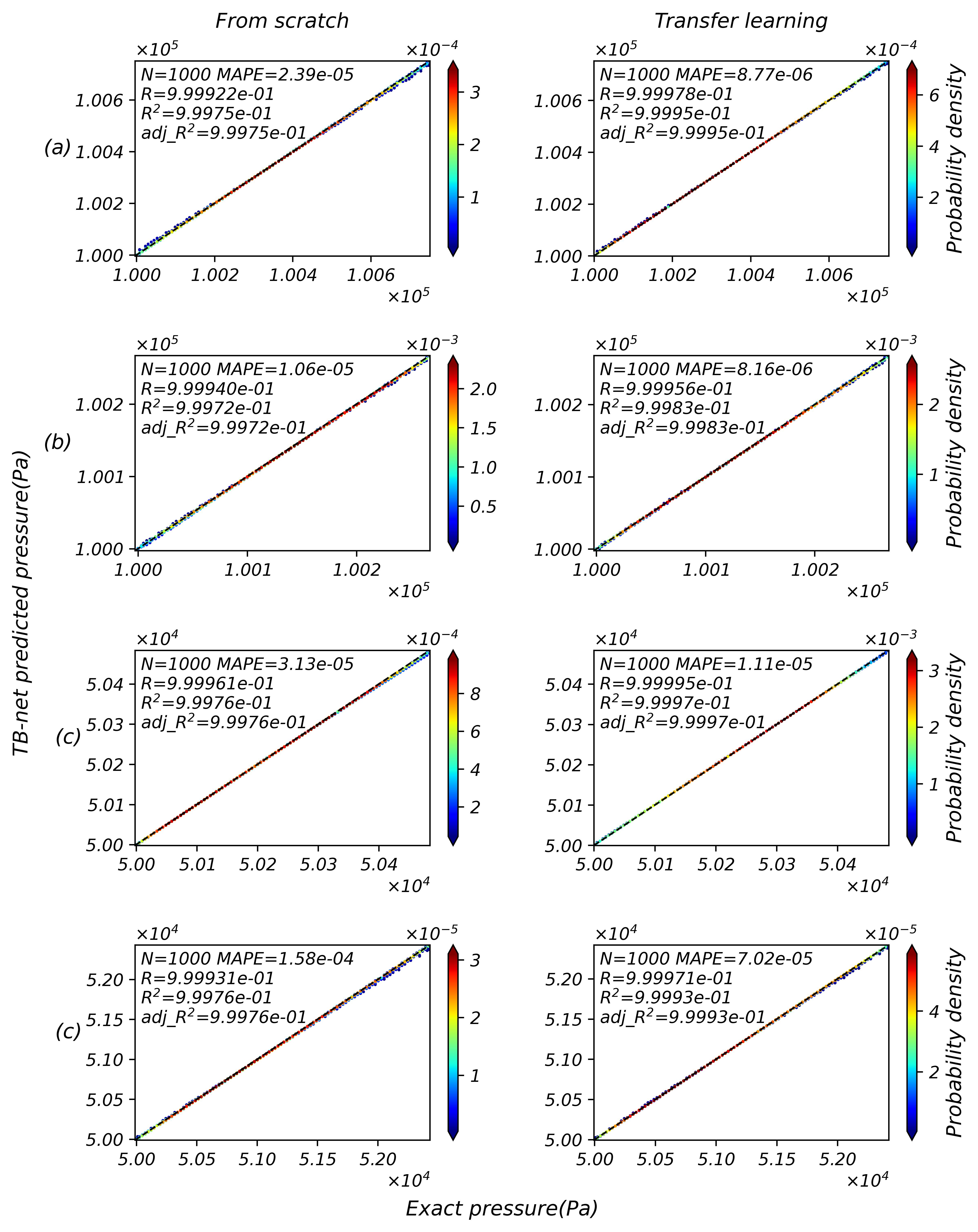}
    \caption{The TB-net predicted and exact pressure distribution and counterpart probability density: (a) Flow problem with $\varepsilon$ = 0.4. The relative $\mathcal{L}_2$ error for model derived by training from scratch and transfer learning is 3.447 $\times$ $10^{-5}$ and 1.610 $\times$ $10^{-5}$, respectively. (b) Flow problem with $\varepsilon$ = 0.5. The relative $\mathcal{L}_2$ error is 1.297 $\times$ $10^{-5}$ and 1.037 $\times$ $10^{-5}$, respectively. (c) Flow problem with \textit{P} = 5.0 $\times$ $10^4$ and $\dot m$ = 0.1. The relative $\mathcal{L}_2$ error is 4.301 $\times$ $10^{-5}$ and 1.459 $\times$ $10^{-5}$, respectively. (d) Flow problem with \textit{P} = 5.0 $\times$ $10^4$ and $\dot m$ = 0.5. The relative $\mathcal{L}_2$ error is 2.236 $\times$ $10^{-4}$ and 1.168 $\times$ $10^{-4}$, respectively.}\label{fig16}
\end{figure}

Fig. \ref{fig15} shows the loss curves corresponding to TB-net PINN from scratch and transfer learning, respectively. When training from scratch, the iteration epoch is set as 1 $\times$ $10^5$, and when it comes to transfer learning, 4 $\times$ $10^4$ is applied. Regardless of whether $\varepsilon$ = 0.4 or $\varepsilon$ = 0.5, the transfer learning procedure exhibits a notably lower loss value from the outset than that of training from scratch, and the loss value rapidly declines and stabilizes within a few epochs. It is evident from the training process that the transfer leaning model possesses an initial state that is closer to the optimization target, and due to the consistency of high-level features, its convergence progresses at a quicker and stable pace. Note that training from scratch involves parameter optimization for both trunk and branch nets, while transfer learning only focus on the branch nets. Therefore, even with equivalent training epochs, transfer learning achieves faster speeds and consumes less memory owing to the diminished AD computations. Additionally, the process for transfer learning is expedited further by the quicker convergence, which necessitates fewer epochs. Theoretically, the training time for case $\varepsilon$=0.4 or $\varepsilon$=0.5 within the same mode should be consistent, and the differences shown in Table \ref{tbl6} are mainly due to the hardware factors and the fine optimization by L-BFGS. Evidently, the adoption of transfer learning results in a roughly twofold increase in speed compared to training from scratch. If operating under the same training mode, training time and epochs are linearly related. Consequently, leveraging the characteristics of the TB-net, solely training the branch nets can save approximately one-tenth of the training time. Apart from enhancing training speed, transfer learning model also maintain a high level of prediction accuracy. As shown in Fig. \ref{fig16}, the predictive performance of the models obtained under different modes is demonstrated through KDE and various statistical metrics. The predicted and exact pressure distributions are almost entirely aligned with the 1:1 reference line, and in terms of quantitative metrics, the prediction performance of transfer learning slightly exceed that achieved through training from scratch.

\begin{table*}[pos=!htbp,width=0.7\textwidth]
\caption{The training time of two modes on different cases shown in transfer learning part. The training is implemented on GeForce RTX 3090 GPU with 10496 CUDA cores.}\label{tbl6}
\begin{tabular*}{\tblwidth}{@{}LCC@{}}
\toprule
   \textbf{Case} & From scratch mode & Transfer learning mode \\ 
\midrule
$\varepsilon$ = 0.4 & 2539.52 s & 909.55 s \\
$\varepsilon$ = 0.5 & 2524.69 s & 911.49 s \\
 $\dot m$ = 0.1 & 2567.64 s & 892.82 s \\
 $\dot m$ = 0.5 & 2555.40 s & 912.31 s \\
linear $\dot q$  & 9278.35 s & 5521.00 s \\
\bottomrule
\end{tabular*}
\end{table*}

\subsubsection{Effect of pressure on the flow field}\label{Section 3.3.2}
Our discussions have centered on flow and heat transfer problems with standard atmosphere pressure on the outlet boundary, yet low-pressure environments are frequently encountered in the realms of aerospace, fuel cell technology, and geometrical systems. Consequently, in this part, building upon the flow problems outlined in case A and case B, the pressure boundary is altered to a low-pressure $p_{Outlet}$ = 5.0 $\times$ $10^4$ Pa, while maintaining all other conditions identical. Except for the characteristic pressure \textit{P} = 5.0 $\times$ $10^4$ Pa, no other characteristic parameters undergo any changes. The OOM of the loss terms $e_1$ $\sim$ $e_3$, $e_6$ $\sim$ $e_7$, $e_{10}$ and $e_{13}$ $\sim$ $e_{14}$ remains unaffected, thus preserving the same weights associated with them. Similar to Section \ref{Section 3.3.1}, comparative computations are performed, and the related training procedure is illustrated in Fig. \ref{fig15}. The transfer learning processes are proceeded by trained models from \textit{case} A and \textit{case} B, respectively, and the corresponding loss curves experience sharp declines within a few epochs and stabilize thereafter. The training time of different training modes are shown in Table \ref{tbl6}, and transfer learning still exhibits a significant increase in speed. Note that the architecture of TB-net and the components of loss function are identical in the first four cases in Table \ref{tbl6}, and the primary reason for the time discrepancies using the same training mode lies in hardware fluctuations and further optimization of the L-BFGS optimizer.

KDE is also conducted using 1000 randomly selected spatial points, and the related distributions and statistical indicators are manifested in Fig. \ref{fig16}. When $\dot m$ = 0.1 kg/($\mathrm{m^2}$s), the relative $\mathcal{L}_2$ error for TB-net training from scratch and transfer learning is 4.301 $\times$ $10^{-5}$ and 1.459 $\times$ $10^{-5}$, respectively. When it comes to $\dot m$ = 0.5 kg/($\mathrm{m^2}$s), the relative $\mathcal{L}_2$ error is 2.236 $\times$ $10^{-4}$ and 1.168 $\times$ $10^{-4}$, respectively. In terms of comprehensive evaluation, the model derived from transfer learning proves to be more effective than one that is trained from scratch.

\subsubsection{Effect of boundary heat flux on the heat transfer}\label{Section 3.3.3}
\begin{figure}[htbp]
  \centering
    \includegraphics[scale=0.6]{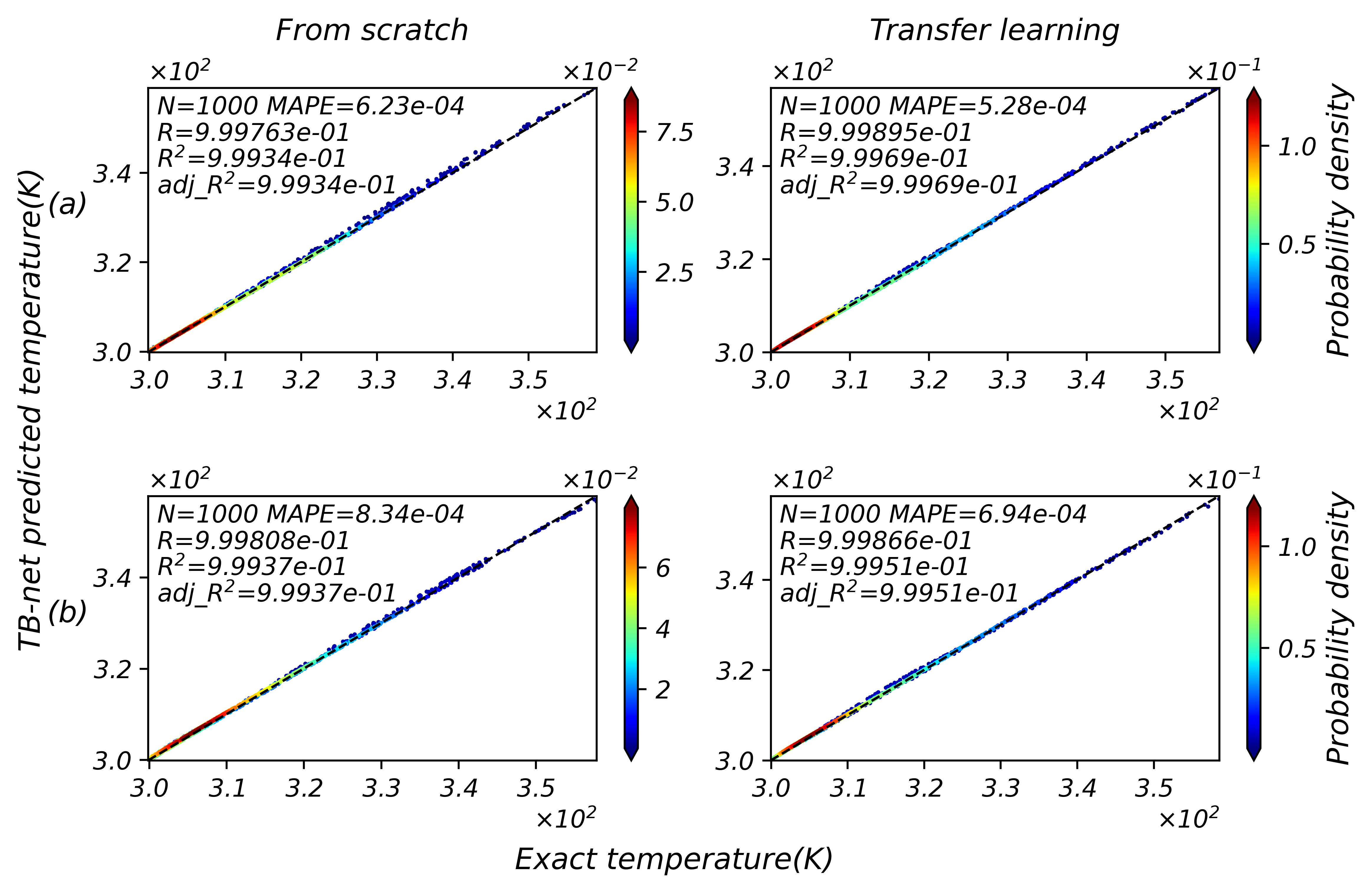}
    \caption{Inverse heat transfer problem with linear $\dot q$: (a) The TB-net predicted and exact $T_s$ distribution and counterpart probability density. The relative $\mathcal{L}_2$ error for TB-net model trained from scratch and transfer learning is 1.037 $\times$ $10^{-3}$ and 7.072 $\times$ $10^{-4}$, respectively. (b) The TB-net predicted and exact $T_f$ distribution and counterpart probability density. The relative $\mathcal{L}_2$ error is 1.056 $\times$ $10^{-3}$ and 9.089 $\times$ $10^{-4}$, respectively.}\label{fig17}
\end{figure}

In Section \ref{Section 3.1.2} and Section \ref{Section 3.2}, forward and inverse heat transfer problems with heat flux boundary set as a constant value have been already studied, however, when dealing with porous heat transfer in chips or for thermal protection of aircraft, it is common to encounter heat flux boundaries that vary in a linear or even nonlinear relation. Here, to showcase the effectiveness of transfer learning and underscore the capability of the TB-net, the inverse heat transfer problem in porous media with linear heat flux boundary condition is investigated:
\begin{equation}
\dot q=(1.5-\widetilde{x})\times 10^5\quad \mathrm{W/m^2} \label{Eq.12}
\end{equation}
which represents the linear relation between non-dimensional horizontal coordinate and the heat flux condition. The other condition is identical to those of the inverse problem for case E, and the number of extra labeled data points on the outlet boundary is set to 20. Considering the variability of the heat flux boundary within a predetermined range, $\lambda_{11}$ is set as $10^{-10}$ for the sake of convenience. The TB-net here has the same architecture and training parameters with that employed in Section \ref{Section 3.2}, and the training epochs for training from scratch and transfer learning are 1 $\times$ $10^5$ and 6 $\times$ $10^4$, respectively. In fact, the initial state of transfer learning mode could be chosen from any trained model of inverse problems with respect to \textit{case} D, \textit{case} E and \textit{case} F, as they all have a connection with the current problem. The trained model of inverse problem pertaining to case E is opted for in this example. No matter whether it is trained from scratch or transfer learning, only the branch nets pertaining to $\widetilde{h_k}$ and $\widetilde{T_s}$ need to be trained, and the relevant training time is displayed in Table \ref{tbl6}. As shown in Fig. \ref{fig15}, the loss curve of transfer learning performs superior than that of training from scratch. Analogously, the KDE and correlation analysis are conducted using the predicted and exact $T_s$ and $T_f$. Fig. \ref{fig17} indicates that training from scratch with TB-net is capable of handling linear heat flux boundary problems, while leveraging transfer learning can improve prediction performance.

In a nutshell, the transfer learning mode starts with pre-existing high-dimensional features, enabling it to be closer to the optimal solution initially and facilitating quicker convergence during the training process. Additionally, the integration of latent features from diverse operating scenarios enhances the precision of the prediction results compared to training from scratch.
A more detailed discussion on the boundaries of transfer learning is presented in the \ref{Appendix F}.

\section{Conclusions and outlook}\label{Section 4}
A novel TB-net PINN architecture is developed in this work to predict the flow and heat transfer in porous medium considering LTNE effect, which is capable of capturing both global and local features to enhance the convergence of training and the accuracy of prediction. A strategy that integrates TB-net PINN with a step-wise training approach is used to address heat transfer problems, including forward problems without labeled data and inverse problems. The main conclusions of this work can be summarized as follows:

\begin{enumerate}[1)]
    \item The TB-net PINN has been proven more efficient, effective and flexible than vanilla FNN configuration in the process of tackling forward flow and heat transfer, inverse heat transfer, and transfer learning problems. Additionally, the OOM and importance analysis for loss term weight design plays a vital role when dealing with such intricate issues.
    \item A systematic study of inverse heat transfer problem reveals the rule regarding how the quantity of boundary labeled temperature data affects the prediction precision. Specifically, even a modest number of labeled data can drastically improve accuracy, offering crucial insights into the combination of experiments and numerical investigations.
    \item Transfer learning techniques can effectively enhance training efficiency and prediction accuracy, while the reutilization of existing models also conserves computational resources.
\end{enumerate}

It is worth mentioning that despite the strengths of PINNs-based methods in being mesh-free, smoothly tackling inverse problems, and utilizing pre-existing models, traditional numerical methods such as FVM and FEM are still the gold standard for solving forward problems at present. However, with the rapid progress in hardware technologies including GPUs and tensor processing units (TPUs), and the further improvement of industrial-grade platforms like PyTorch and TensroFlow, as well as the continuous research into NN architecture design (such as hybrid frameworks with deep operator network and Fourier neural operator), loss function combinations and advanced optimization methods, it is promising that PINNs-like methods could serve as an alternative approach for addressing forward problems in the near future.

This work shows such an example in exploring flow and heat transfer problems in porous media, encompassing both forward and inverse problems as well as transfer learning perspective. However, as porosity is treated as a constant in these problems, resulting in rather simplistic flow characteristics, further studies with graded porosity porous media are expected subsequently. In addition, only single-phase flow is considered in the current work. The phase-change problems could be explored further, under both steady and transient conditions.

\section*{Declaration of competing interest}
The authors declare that they have no known competing financial interests or personal relationships that could have
appeared to influence the work reported in this paper.

\section*{Acknowledgement}
This research is supported by the project of National Natural Science Foundation of China (No. 52206067), NSFC-DFG Mobility Programme (No. M-0368) and 111 Centre (B18002).

\appendix
\renewcommand{\appendixname}{Appendix}
\renewcommand{\thesection}{\appendixname~\Alph{section}}
\section{Sensitivity analysis for the weight parameters}\label{Appendix A}
\setcounter{equation}{0}
\renewcommand{\theequation}{A.\arabic{equation}}
\setcounter{figure}{0}
\renewcommand{\thefigure}{A.\arabic{figure}}
\setcounter{table}{0}
\renewcommand{\thetable}{A\arabic{table}}
The gradient descent method is a crucial approach currently employed for neural network training. 
It obtains the specific update information for each parameter by computing the gradient for the loss function. Consequently, the update status of the parameter set $\sigma$ can be regarded as
\begin{equation}
\sigma_{new}=\sigma_{old}-\eta\nabla_{\sigma}\mathcal{L}_{Data}-\eta\nabla_{\sigma}\mathcal{L}_{PDE} - \eta\nabla_{\sigma}\mathcal{L}_{Inlet} - \eta\nabla_{\sigma}\mathcal{L}_{Outlet}-\eta\nabla_{\sigma}\mathcal{L}_{Wall}, \label{Eq.A.1}
\end{equation}
where $\sigma_{old}$ represents the parameter set before updating, $\sigma_{new}$ denotes the parameter set after the update, $\eta$ is the learning rate, and $\nabla$ is the differential operator used for calculating gradients. As can be seen, the updating of neural network parameters is highly correlated with the gradients of the individual terms in the loss function. When there is an imbalance between the various gradient components, it is prone to resulting in a non-convergent outcome.
In our work, using the normalized order-of-magnitudes (OOM) of the loss components as the weights, approximating gradients with the magnitudes. Following that, specific weights are fine-tuned based on an importance analysis. As for why the value of $\lambda_6$ is revised to $10^2$ in forward problems, it is because if the value is excessively small, it will be unable to adequately fit the inlet velocity. On the other hand, if the value is overly large, it will hinder the training of other loss terms.
\begin{figure}[htbp]
  \centering
    \includegraphics[scale=0.8]{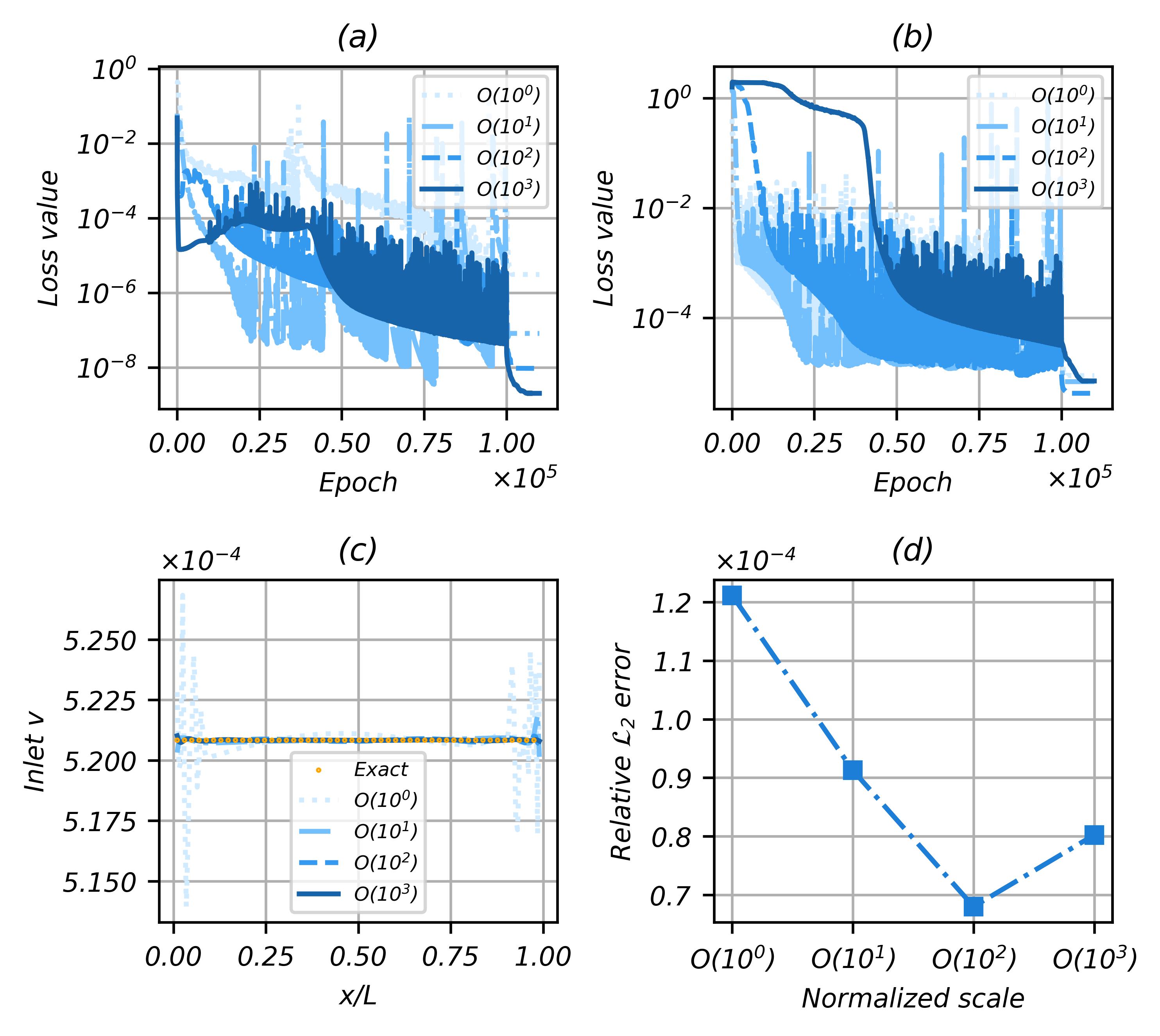}
    \caption{Flow problem with $\dot{m}$ = 0.5 kg/$\mathrm{m^2}s$: (a) The history of $e_6$, (b) the history of $\mathcal{L}_\sigma-\lambda_6e_6$, (c) the inlet $v$ prediction results, and (d) the relative $\mathcal{L}_2$ error of predictive $p$ under different normalized scales.}\label{figA1}
\end{figure}
\begin{table*}[pos=!htbp,width=0.7\textwidth]
  \caption{The final $e_6$, $\mathcal{L}_\sigma-\lambda_6e_6$, relative $\mathcal{L}_2$ error for predictive inlet $v$, and relative $\mathcal{L}_2$ error for predictive $p$ under different normalized scales.}\label{tblA1}
  \begin{tabular*}{\tblwidth}{@{}LLLLL@{}}
  \toprule
   \textbf{Normalized scale} & $e_6$ & $\mathcal{L}_\sigma-\lambda_6e_6$ & relative $\mathcal{L}_2$ - inlet $v$ & relative $\mathcal{L}_2$ - $p$ \\ 
  \midrule
   $O$($10^0$) & 3.06 $\times$ $10^{-6}$ & 8.82 $\times$ $10^{-6}$ & 2.09 $\times$ $10^{-3}$ & 1.21 $\times$ $10^{-4}$ \\
   $O$($10^1$) & 8.08 $\times$ $10^{-8}$ & 6.86 $\times$ $10^{-6}$ & 2.33 $\times$ $10^{-4}$ & 9.13 $\times$ $10^{-5}$ \\
   $O$($10^2$) & 9.37 $\times$ $10^{-9}$ & 4.21 $\times$ $10^{-6}$ & 9.43 $\times$ $10^{-5}$ & 7.07 $\times$ $10^{-5}$ \\
   $O$($10^3$) & 2.01 $\times$ $10^{-9}$ & 7.07 $\times$ $10^{-6}$ & 4.15 $\times$ $10^{-5}$ & 8.02 $\times$ $10^{-5}$ \\
  \bottomrule
  \end{tabular*}
  \end{table*}

To provide a more specific explanation, a systematic sensitivity analysis is conducted. $\lambda_6$ in our work is set to $10^0$, $10^1$, $10^2$, and $10^3$, respectively, corresponding to the OOM of \textit{O}($10^0$), \textit{O}($10^1$), \textit{O}($10^2$), and \textit{O}($10^3$) for the inlet $v$ loss term $e_6$ in our work. 
For the sake of intuitive comparison, experiments were conducted on \textit{case} B using the Adams optimizer for 1 $\times$ $10^5$ steps and the L-BFGS optimizer for 1 $\times$ $10^4$ steps. Fig. \ref{figA1}(a) shows the loss term value $e_6$ pertaining to inlet $v$, and as the normalized scale increases, the value of $e_6$ decreases, leading to a better fitting performance for the inlet $v$. Nevertheless, the curves shown in Fig. \ref{figA1}(b) indicate that other loss $\mathcal{L}_\sigma-\lambda_6e_6$ with normalized scale $O$($10^2$) are trained more thoroughly. This means that extremely large weight $\lambda_6$ has affected the training of other loss terms. The corresponding quantitative analyses are shown in Table \ref{tblA1}. It can be observed that the final full field pressure prediction result obtained by normalized scale $O$($10^2$) exhibits the best performance. 

\section{Trend analysis of predictive error disparities between the TB-net and FNN frameworks}\label{Appendix B}
\setcounter{equation}{0}
\renewcommand{\theequation}{B.\arabic{equation}}
\setcounter{figure}{0}
\renewcommand{\thefigure}{B.\arabic{figure}}
\setcounter{table}{0}
\renewcommand{\thetable}{B\arabic{table}}
To investigate the error trends in prediction results from the two frameworks, i.e., TB-net and FNN, we have incorporated an additional case $\dot{m}=2.0$ kg/($m^2$s) in this part.
The selection of collocation points and other training hyperparameters are consistent with those in $case$ A, and the final quantitative results are presented in Fig. \ref{figB1} and Table \ref{tblB1}.
It seems that the gap in relative $\mathcal{L}_2$ error is narrowing as the mass flux increases, and there is even a tendency for the accuracy of FNN to surpass that of TB-net.
However, the main reason is that with an increase in mass flux, the pressure drop grows, and the absolute pressure magnitude climbs. Moreover, a wider distribution range of variables makes prediction more challenging. In essence, TB-net still consistently outperforms FNN. 
\begin{figure}[htbp]
  \centering
    \includegraphics[scale=0.8]{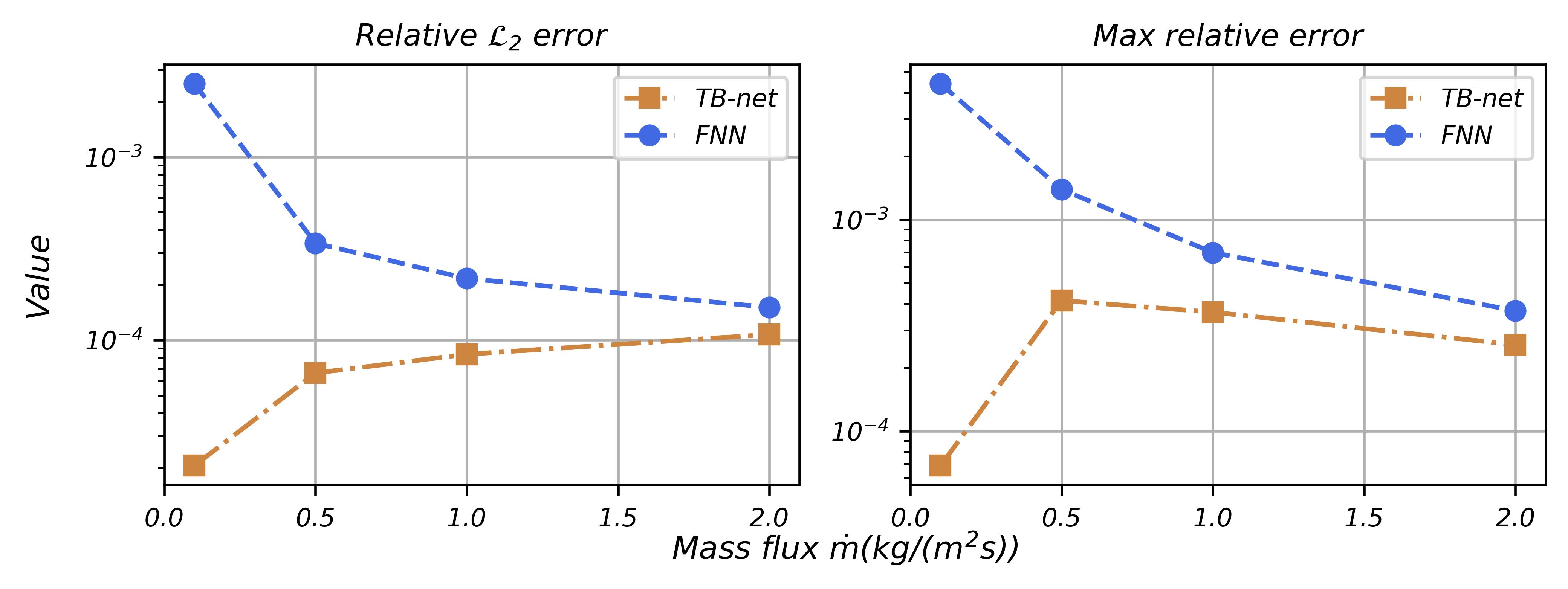}
    \caption{ The relative $\mathcal{L}_2$ and max error of pressure predicted from TB-net and FNN with respect to the exact pressure.}\label{figB1}
\end{figure}
\begin{table*}[pos=!htbp,width=0.7\textwidth]
  \caption{The relative $\mathcal{L}_2$ and max relative error of pressure predicted from TB-net and FNN with respect to the exact pressure under different mass fluxes.}\label{tblB1}
  \begin{tabular*}{\tblwidth}{@{}LLLLL@{}}
  \toprule
     & \multicolumn{4}{c}{TB-net} \\ 
  \midrule
   $\dot{m}/(kg/(m^2s))$ & 0.1 & 0.5 & 1.0 & 2.0 \\
  \midrule
    Relative $\mathcal{L}_2$ error & 2.06 $\times$ $10^{-5}$ & 6.62 $\times$ $10^{-5}$ & 8.37 $\times$ $10^{-5}$ & 1.07 $\times$ $10^{-4}$ \\
     Max relative error & 6.86 $\times$ $10^{-5}$ & 4.15 $\times$ $10^{-4}$ & 3.65 $\times$ $10^{-4}$ & 2.56 $\times$ $10^{-4}$ \\
  \midrule
  & \multicolumn{4}{c}{FNN} \\
  \midrule
   $\dot{m}/(kg/(m^2s))$ & 0.1 & 0.5 & 1.0 & 2.0 \\
   \midrule
  Relative $\mathcal{L}_2$ error & 2.52 $\times$ $10^{-3}$ & 3.38 $\times$ $10^{-4}$ & 2.17 $\times$ $10^{-4}$ & 1.51 $\times$ $10^{-4}$ \\
  Max relative error & 4.41 $\times$ $10^{-3}$ & 1.39 $\times$ $10^{-3}$ & 6.99 $\times$ $10^{-4}$ & 3.71 $\times$ $10^{-4}$ \\
  \bottomrule
  \end{tabular*}
  \end{table*}
\section{Exploration into the three-dimensional porous flows}\label{Appendix C}
\setcounter{equation}{0}
\renewcommand{\theequation}{C.\arabic{equation}}
\setcounter{figure}{0}
\renewcommand{\thefigure}{C.\arabic{figure}}
When switching from a two-dimensional case to a three-dimensional one, the corresponding governing equations will become three-dimensional. Considering the physical model depicted in Fig. \ref{figC1}, the four boundary surfaces perpendicular to the xOz plane are no-slip walls. The mass flux inlet and pressure outlet are employed. Here, the characteristic length $L$ = 0.02 m, and two selected planes $\widetilde{x}$ = 0.3 and $\widetilde{x}$ = 0.7 are utilized for the purpose of verifying the predicted results. The mathematical model for the incompressible, viscid flow in the porous media can be formulated as:
\begin{subequations}
  \begin{align}
          &u_x + v_y + w_z = 0, \label{Eq.C1a}& \\
          &\frac{\rho}{\varepsilon}((u^2)_x+(uv)_y+(uw)_z)=-p_x-\frac{\mu}{K}u, \label{Eq.C1b}& \\
          &\frac{\rho}{\varepsilon}((vu)_x+(v^2)_y+(vw)_z)=-p_y-\frac{\mu}{K}v, \label{Eq.C1c}& \\
          &\frac{\rho}{\varepsilon}((uw)_x+(vw)_y+(w^2)_z)=-p_z-\frac{\mu}{K}w, \label{Eq.C1d}&
      \end{align}
  \end{subequations}
where $w$ denotes the velocity of the liquid along the z-axis. The non-dimensional form of Eqs.(\ref{Eq.C1a}-\ref{Eq.C1d}) will be embedded into the $\mathcal{L}_{PDE}$. Meanwhile, the other associated $\mathcal{L}_{Inlet}$, $\mathcal{L}_{Outlet}$ and $\mathcal{L}_{Wall}$ will also be rewritten in a three-dimensional form, which means the overall count of loss terms will rise even more. Here, the mass flux $\dot{m}$ = 0.5 kg/($\mathrm{m^2}$s) and the pressure at the outlet is $10^5$ Pa. The corresponding characteristic velocity $V$ = 0.5/$\rho$, and the characteristic pressure $P$ = ${10}^5$ Pa.
\begin{figure}[htbp]
  \centering
    \includegraphics[scale=0.35]{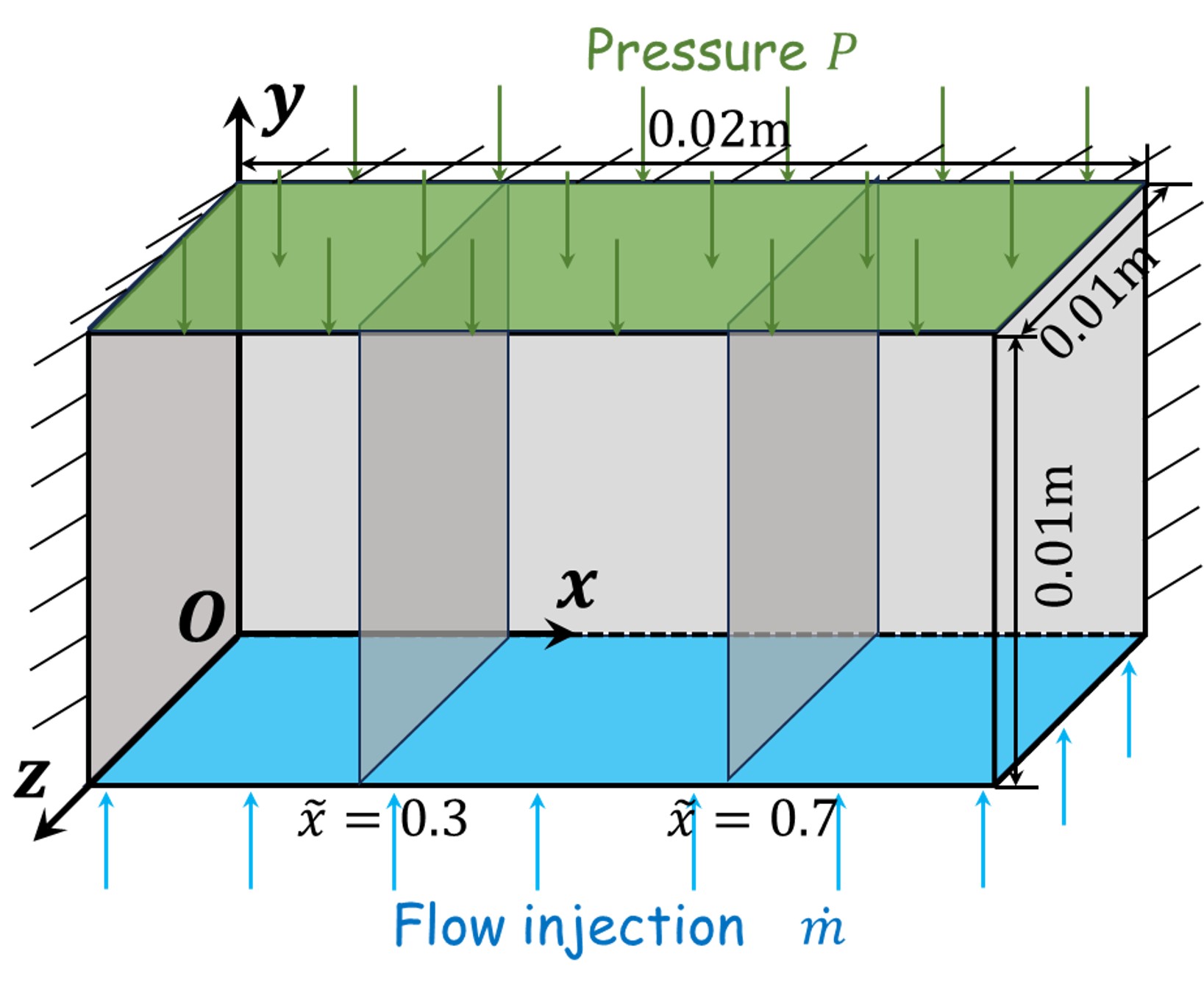}
    \caption{The three-dimensional physical model of the flow in the porous medium employed in this work.}\label{figC1}
\end{figure}
\begin{figure}[htbp]
  \centering
    \includegraphics[scale=0.55]{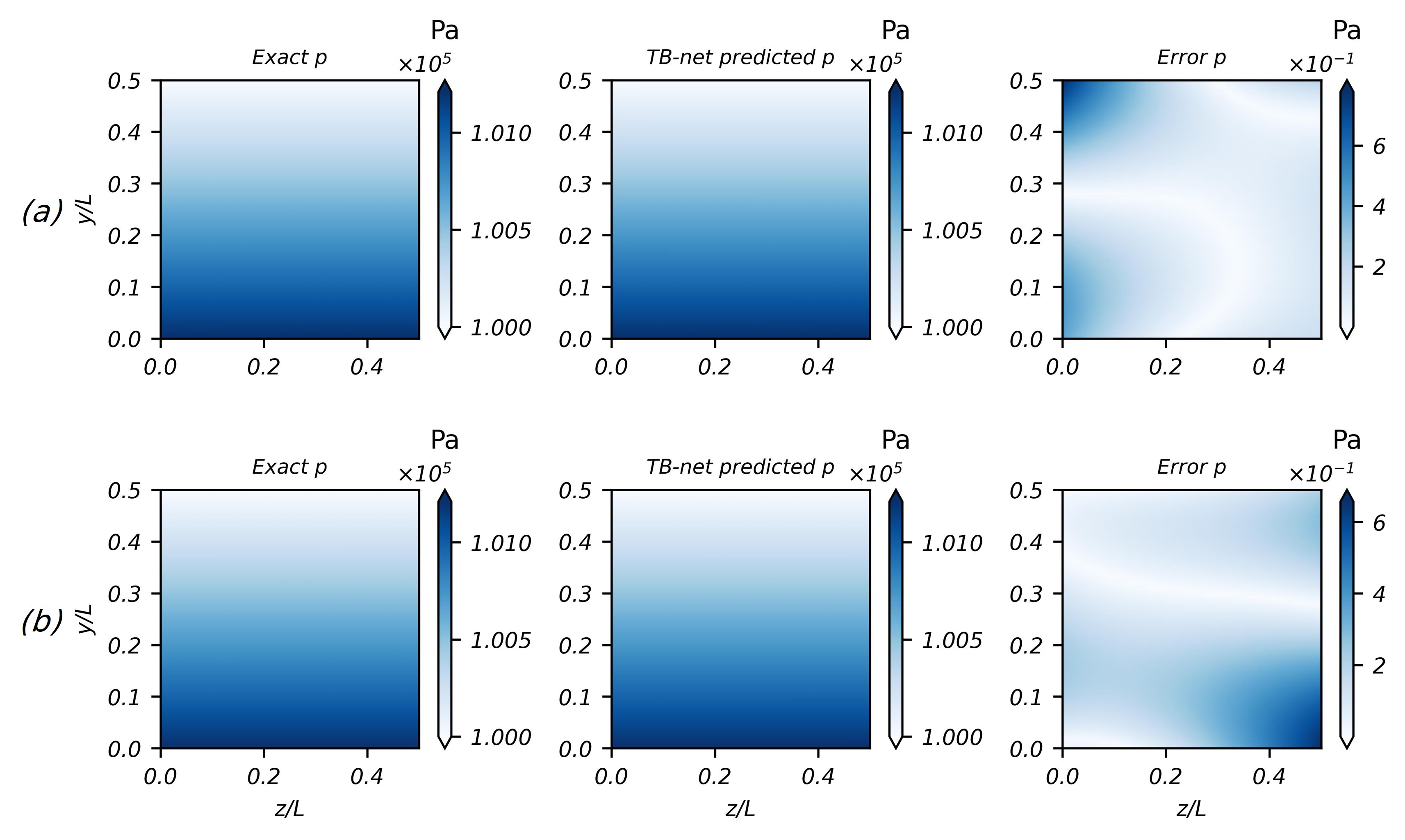}
    \caption{Three-dimensional flow problem with $\dot{m}$ = 0.5 kg/($\mathrm{m^2}s$): (a): Comparison of exact and TB-net predicted pressure on plane $\widetilde{x}$ = 0.3. (b): Comparison of exact and TB-net predicted pressure on plane $\widetilde{x}$ = 0.7.}\label{figC2}
\end{figure}

The inputs of the network contain $\widetilde{x}$, $\widetilde{y}$ and $\widetilde{z}$, where $\widetilde{z}$ = $z/L$. As the increase in the problem's dimensionality leads to an expansion in the mapping space, the TB-net structure is set as 5-layer trunk net with 150 neurons per layer, and 3-layer deep branch nets with 50 neurons per layer. The training set comprises $N_{Inlet}$ = 5000, $N_{Outlet}$ = 5000, and $N_{Wall}$ = 4000 for four surfaces to enforce the boundary conditions, and $N_{PDE}$ = 100000 for ensuring the PDEs in the computational domain. The collocation points are generated through LHS (Latin hypercube sampling). First the Adams optimizer is employed to train the model for 1 $\times 10^5$ epochs with a learning rate of 1$e$-4.
Subsequently, the L-BFGS optimizer is used for fine-tuning. In order to verify the accuracy of the predictive results, two planes $\widetilde{x}$ = 0.3 and $\widetilde{x}$ = 0.7 are selected to quantify the pressure prediction precision, as illustrated in Fig. \ref{figC2}. The relative $\mathcal{L}_2$ error of the predictive pressure with respect to the exact pressure on the plane $\widetilde{x}$ = 0.3 is 1.854 $\times$ $10^{-6}$, and 2.097$\times 10^{-6}$ on the plane $\widetilde{x}$ = 0.7. It can be shown that, in the context of three-dimensional porous flow, the TB-net framework can still maintain high stability and prediction accuracy. It is worth noting that the three-dimensional case here is rather analogous to case B of the two-dimensional case regarding boundary conditions.
However, due to the increased complexity of the problem resulting from the elevation of dimensionality, the overall network structure requires more neurons to cover the more complex dynamic processes. Meanwhile, the number of collocation points also needs to increase explosively to fully constrain the boundary conditions and the PDEs. The case presented here actually involves simple flow phenomenon, and it is primarily intended for testing the model's capability to handle three-dimensional cases. When dealing with highly complex flow scenarios, apart from strengthening the model's expressive power, one also need's to carefully design the distribution of the collocation points. Otherwise, key flow details are likely to be overlooked.
Similar to the process of elevating the problem from two-dimensional to three-dimensional, the transition from steady-state to transient conditions also necessitates the addition of a variable in the input layer, namely the dimensionless time $\widetilde{t}$. In addition to this, it is essential to expand the model capacity and increase the number of collocation points. 
\section{Comparison between coupled and step-wise sequential schemes}\label{Appendix D}
\setcounter{equation}{0}
\renewcommand{\theequation}{D.\arabic{equation}}
\setcounter{figure}{0}
\renewcommand{\thefigure}{D.\arabic{figure}}
To investigate the stability and effectiveness of the step-wise sequential training scheme, we conduct a comparison between the sequential scheme and three fully joint training schemes, i.e., joint-1, joint-2 and joint-3. Note that $\mathcal{L}_{flow}$ means the total loss adopted for the flow problem, and $\mathcal{L}_\sigma$ here increases loss items related to heat transfer. The sequential training scheme is performed consistent with that in the paper, on the basis of the pre-trained model in $case$ B, only loss terms pertaining to heat transfer process are involved. 2 $\times 10^4$ epochs with Adams optimizer at a learning rate of 1$e$-4 and 1 $\times 10^4$ epochs with L-BFGS optimizer are employed. In joint-1, loss terms for both flow and heat transfer are grouped together to constitute the loss function.
There is no pre-trained model involved. The training from scratch starts with the Adams optimizer for 5 $\times 10^3$ epochs at a learning rate of 1$e$-4, and then it shifts to the L-BFGS optimizer for 1 $\times 10^4$ epochs. Joint-2 adopts the Adams optimizer for 5 $\times 10^3$ epochs with a learning rate of 5$e$-5, and subsequently switches to the L-BFGS optimizer for 1 $\times 10^4$ epochs. Joint-3 utilizes the Adams optimizer for 2 $\times 10^4$ epochs at a learning rate of 5$e$-5, and then changes to the L-BFGS optimizer for 1 $\times 10^4$ epochs. However, none of the fully joint training schemes could converge without issues. As shown in Fig. \ref{figD1}, all three joint training schemes halted training before 7 $\times 10^3$ epochs (as indicated by the gray dashed lines in Fig. \ref{figD1}(a) and \ref{figD1}(b)), while under ideal circumstances, the training process ought to be halted once the required number of the epochs for each case is achieved.
The reason for the early termination was that gradient explosion had an irreversible impact on the network parameters, rendering all loss values invalid and making it impossible to continue the training. The root cause of the gradient explosion was the exceedingly intricate structure of the loss function, which gave rise to a highly irregular gradient space. In the current setup, this situation was very likely to trigger sharp parameter fluctuations. Specifically, joint-1 diverged at the 4.95 $\times 10^{3th}$ step and failed to reach L-BFGS optimizer stage. Since the learning rate is 1$e$-4, we consider reducing it to 5$e$-5 to mitigate the chance of gradient explosion. Joint-2 completed the Adams stage. However, it diverged directly after only 100 epochs of L-BFGS optimization.
This was actually because the L-BFGS optimizer utilizes second-order derivatives, making it more sensitive to variations in gradients. Joint-3, on the other hand, diverged after being trained with the Adams optimizer for 7 $\times 10^3$ epochs and thus couldn't accomplish the training. Fig. \ref{figD1}(a) essentially depicts the changes in losses related to heat transfer. Prior to the divergence of the fully joint schemes, it doesn't seem to differ much from the sequential scheme. This is due to the fact that the initial loss values are on a large scale and are not difficult to reduce. Nevertheless, the subsequent optimization is the key to distinguishing the capabilities of the models. Fig. \ref{figD1}(b) offers a more straightforward illustration of the gap in flow losses between the fully joint training schemes and the sequential scheme. 
\begin{figure}[htbp]
  \centering
    \includegraphics[scale=1.0]{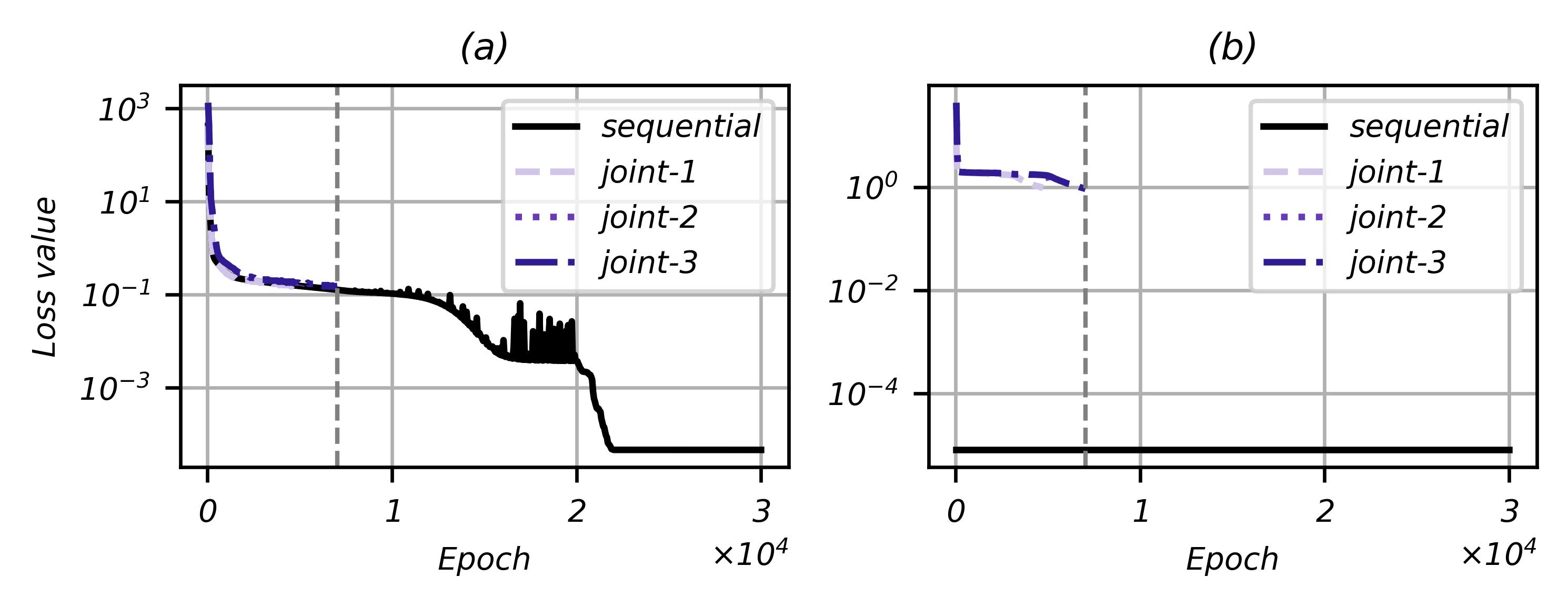}
    \caption{Heat transfer problem with $\dot{q}$ = 5.0 $\times {10}^4$ W/$\mathrm{m^2}$: (a) The history of $\mathcal{L}_\sigma-\mathcal{L}_{flow}$, and (b) the history of $\mathcal{L}_{flow}$ under different training schemes.}\label{figD1}
\end{figure}

In this work, we have introduced reasonable simplifications to the model, i.e. approximately estimate $\mu$ and $Pr$ under reference temperature 300 K, enabling the sequential solution of flow and heat transfer issues. Compared with the joint schemes under the identical conditions, the TB-net step-wise scheme stands out for its stability and effectiveness. Moreover, it can be extended to a series of sequential-solution problems. When dealing with real-world, highly complex without any assumptions, the fully-joint format becomes indispensable. However, this demands a more elaborated designed loss function, learning rate arrangement, and neural network architecture, which is one of the directions we aim to make breakthroughs in the future.
\section{Evaluations for inverse process under noisy conditions}\label{Appendix E}
\setcounter{equation}{0}
\renewcommand{\theequation}{E.\arabic{equation}}
\setcounter{figure}{0}
\renewcommand{\thefigure}{E.\arabic{figure}}
\setcounter{table}{0}
\renewcommand{\thetable}{E\arabic{table}}
In Section \ref{Section 3.2}, relevant research on inverse problems have been carried out, where the numerical solutions were directly employed as experimental data for full-field temperature inversion. However, the data obtained during experimental processes inevitably contain random errors. In this context, we randomly sampled 50 surface labeled data and imposed four different levels of random noise on them, i.e., 0.1\%, 0.2\%, 0.5\%, and 1.0\%.
As a result, the data utilized for solving the inverse problems will be more in line with actual situations. When conducting inverse problem training using surface data that has noise incorporated, the hypermeter configuration is identical to that described in Section \ref{Section 3.2}.

\begin{figure}[htbp]
  \centering
    \includegraphics[scale=0.9]{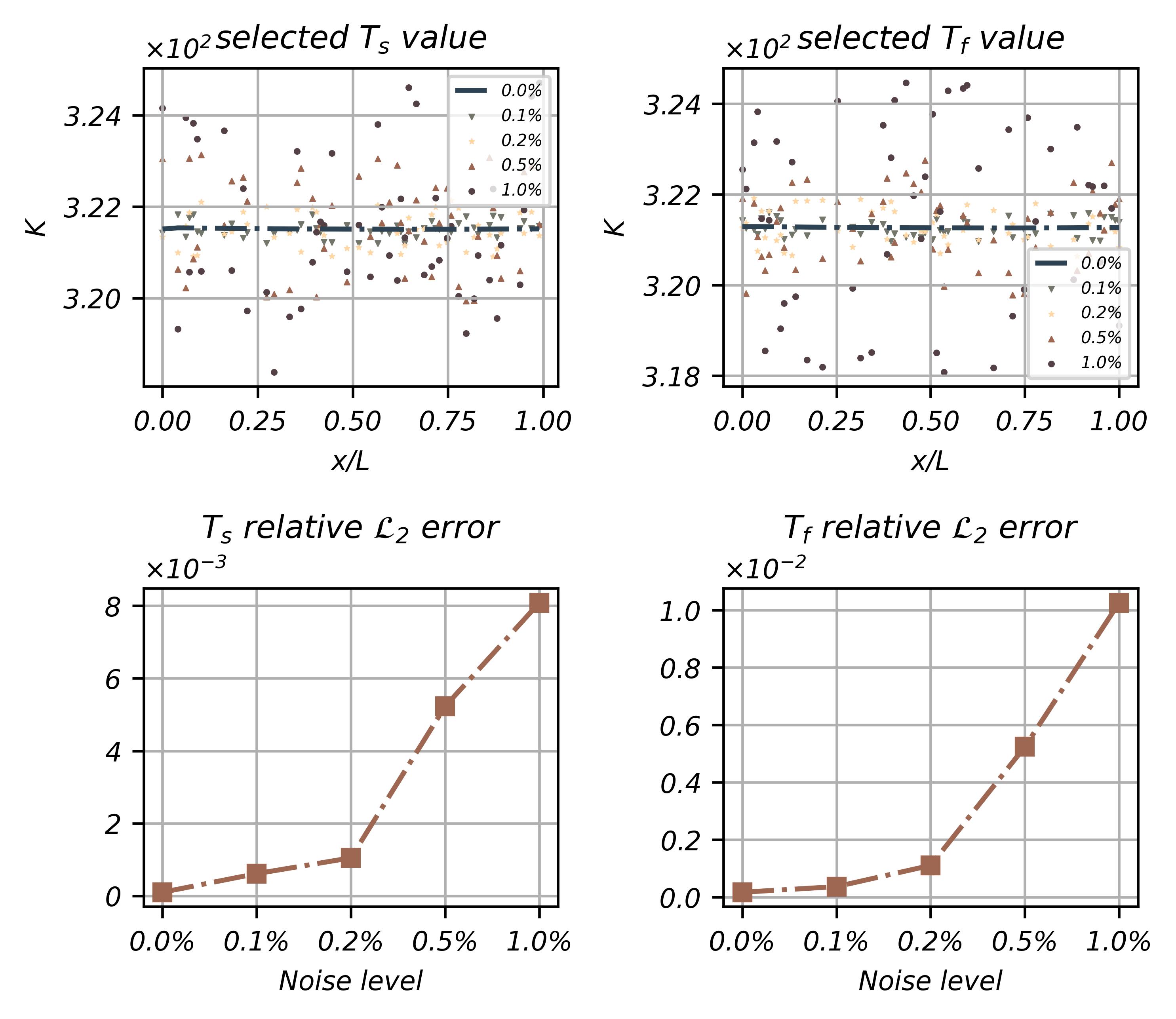}
    \caption{Surface $T_s$ and $T_f$ data with varying levels of noise for inverse problem solving, along with the relative $\mathcal{L}_2$ errors of predictive $T_s$ and $T_f$ under different noise levels.}\label{figE1}
\end{figure}

As illustrated in Fig. \ref{figE1} and Table \ref{tblE1}, with the increase in the noise level of the surface labeled $T_s$ and $T_f$, the relative $\mathcal{L}_2$ errors of the full-field prediction compared to the exact values exhibits considerable rises. The underlying cause is that the labeled data points provide extra constraints, and the deviation of these constraints give rise to the deviation of the full-field prediction results. However, from the perspective pf the magnitude of errors, using noise of a particular level for inverse problem solving can sustain an acceptable degree of prediction accuracy. 
\begin{table*}[pos=!htbp,width=0.7\textwidth]
  \caption{The relative $\mathcal{L}_2$ error for predictive inlet $T_s$, and $T_f$ under different noise levels.}\label{tblE1}
  \begin{tabular*}{\tblwidth}{@{}LLLLLL@{}}
  \toprule
   \textbf{Noise level} & 0.0\% & 0.1\% & 0.2\% & 0.5\% & 1.0\% \\ 
  \midrule
   $T_s$ & 1.73 $\times$ $10^{-4}$ & 6.10 $\times$ $10^{-4}$ & 1.04 $\times$ $10^{-3}$ & 5.22 $\times$ $10^{-3}$ & 8.08 $\times$ $10^{-3}$ \\
   $T_f$ & 2.39 $\times$ $10^{-4}$ & 3.65 $\times$ $10^{-4}$ & 1.10 $\times$ $10^{-3}$ & 5.23 $\times$ $10^{-3}$ & 1.02 $\times$ $10^{-2}$ \\
  \bottomrule
  \end{tabular*}
  \end{table*}
\section{Discussions on the boundary of transfer learning}\label{Appendix F}
\setcounter{equation}{0}
\renewcommand{\theequation}{F.\arabic{equation}}
\setcounter{figure}{0}
\renewcommand{\thefigure}{F.\arabic{figure}}
Transfer learning serves as a means of knowledge reutilization. In the context of this work, its inherent nature is:
\begin{equation}
\sigma_{pre}\rightarrow\sigma_{tar}, \label{Eq.F.1}
\end{equation}
where $\sigma_{pre}$ is the parameter set from the pre-trained model, and $\sigma_{tar}$ denotes the parameter set from the target model. The regular training procedure of a neural network can be regarded as:
\begin{equation}
\sigma_{ini}\rightarrow\sigma_{tar}, \label{Eq.F.2}
\end{equation}
where $\sigma_{ini}$ is the parameter set initialized randomly from the initial model.

Assume that we have already trained a network to deal with a particular problem earlier (the parameter set of this network is denoted as $\sigma_{pre}$), and now, another problem that is basically similar to the previous one while has some differences need to be addressed. Training from scratch would lead to a waste of resources. This is because, through training, $\sigma_{pre}$ already contains certain high-dimensional mapping relationships that are also applicable to the current problem. In the parameter space, the distance between $\sigma_{pre}$ and $\sigma_{tar}$ is much closer compared to the distance between $\sigma_{ini}$ and $\sigma_{tar}$, which means the training will be more efficient and stable through Eq.(\ref{Eq.F.1}). 
\begin{figure}[htbp]
  \centering
    \includegraphics[scale=0.9]{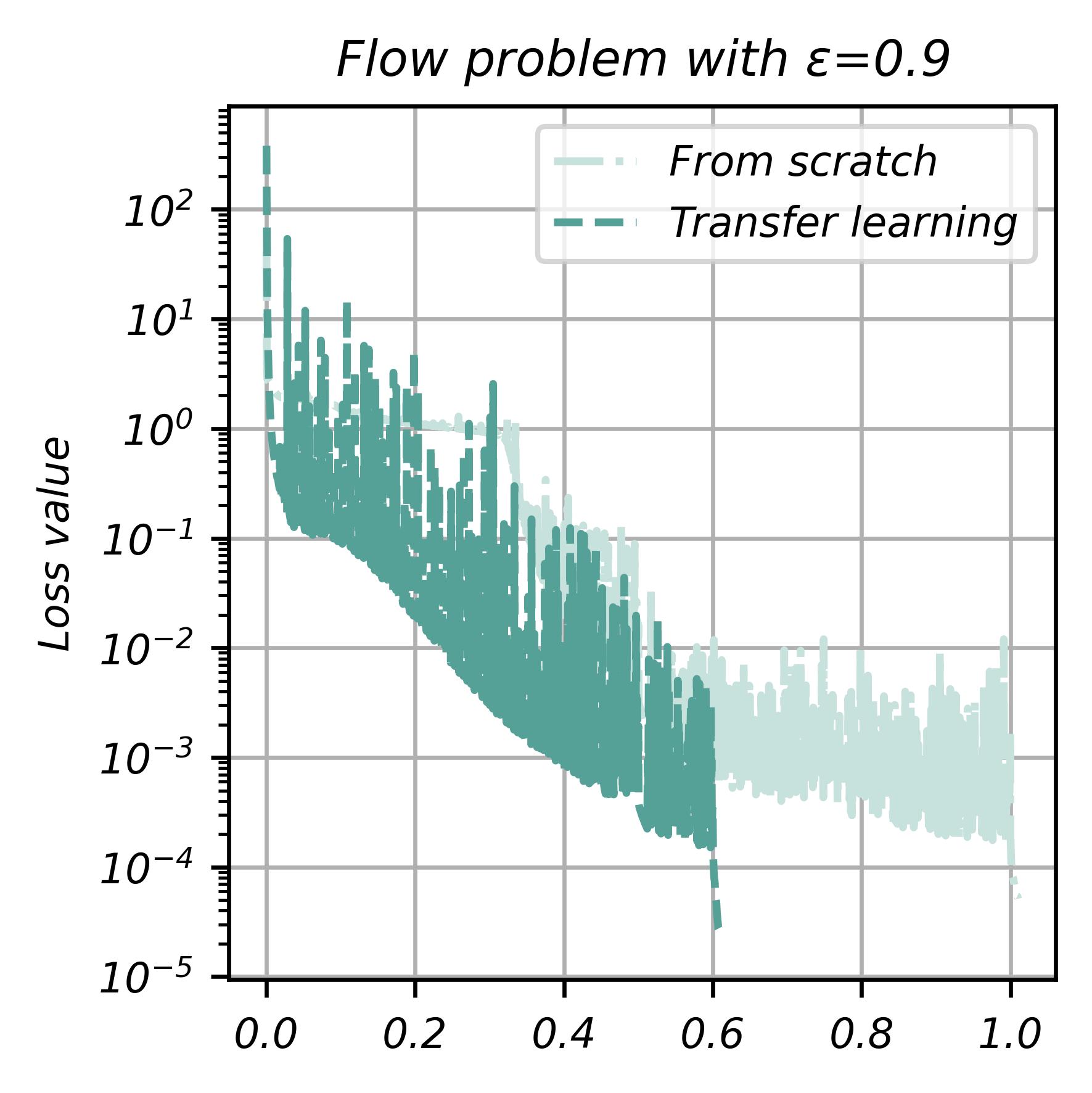}
    \caption{The loss value curves corresponding to TB-net PINN training from scratch and transfer learning.}\label{figF1}
\end{figure}

To further demonstrate the reliability of transfer learning with our framework, the pre-trained flow model at $\varepsilon$ = 0.3 is migrated to a scenario where $\varepsilon$ = 0.9 and K = 4.86 $\times 10^{-9}$. The selection of collocation points remains the same as that in Section \ref{Section 3.3.1}. Here, the number of epochs for the Adams optimizer is set to 6 $\times 10^4$, and then it is switched to the L-BFGS optimizer. As can be observed from Fig. \ref{figF1}, the loss curve of the transfer learning process based on the pre-trained model converges faster compared to training from scratch. The relative $\mathcal{L}_2$ error for the resulting pressure prediction of model trained from scratch is 4.921 $\times 10^{-4}$, and 1.370 $\times 10^{-4}$ for transfer learning model. In this case, transfer learning technique still holds its effectiveness. This is because the boundary conditions for the cases where $\varepsilon$ = 0.3 and $\varepsilon$ = 0.9 are consistent, and the essential flow behaviors share a particular level of resemblance.
Meanwhile, when comparing the curves in Fig. \ref{figF1} with that in Fig. \ref{fig15}, it is evident that the convergence of transfer learning becomes significantly slower. This is because, although the nature of these flows is the same, it is easier to transfer from $\varepsilon$ = 0.3 to $\varepsilon$ = 0.4 or $\varepsilon$ = 0.5 than to $\varepsilon$ = 0.9. In terms of the parameter space, the parameter set $\sigma_{pre}$ obtained from the $\varepsilon$ = 0.3 flow problem is closer to the target parameter set $\sigma_{tar}$ of $\varepsilon$ = 0.4 or $\varepsilon$ = 0.5 flow problem than to that of $\varepsilon$ = 0.9. 
In general, the application of transfer learning is based on the premise that the pre-trained model is closer to the target model in the parameter space than an initialized model is. This technique is commonly adopted for problems exhibiting notable similarities in dynamic processes of for dimensionality-reduction transfer from intricate processes to simple ones. That is, $\sigma_{pre}$ is closer to $\sigma_{tar}$ than $\sigma_{ini}$ is.


\end{document}